%% file: main.tex
\documentclass{egpubl}
\usepackage{eg2027}

\ConferenceSubmission

\usepackage[T1]{fontenc}
\usepackage{dfadobe}

\usepackage{cite}  %
\BibtexOrBiblatex

\electronicVersion
\PrintedOrElectronic

\usepackage{graphicx}

\usepackage{egweblnk}

\usepackage{dblfloatfix}

\input{macros}

\title[HimmBO: High-Dimensional Interactive Model-Merging Bayesian Optimization for Image Generation]%
      {HimmBO: High-Dimensional Interactive Model-Merging Bayesian Optimization for Image Generation}

\author[J. Mao, Z. Shen \& C. Liu]{Jinyuan Mao$^{1}$, Zhihao Shen$^{2}$, and Chenxi Liu$^{3}$\thanks{Corresponding author}\\
{\small $^{1}$Zhejiang University, \quad $^{2}$pix.ai, \quad $^{3}$University of Victoria}}

\makeatletter
\copyrightTextTitPag{}
\copyrightTextRunPag{}
\def\ps@titlepage{\let\@mkboth\@gobbletwo
  \def\@oddhead{}
  \def\@oddfoot{}
  \let\@evenhead=\@oddhead
  \let\@evenfoot=\@oddfoot
  \let\sectionmark=\EmptySectionmark
  \let\subsectionmark=\EmptySubsectionmark
}
\makeatother

\begin{document}

\teaser{%
  \centering
  \includegraphics[width=\linewidth]{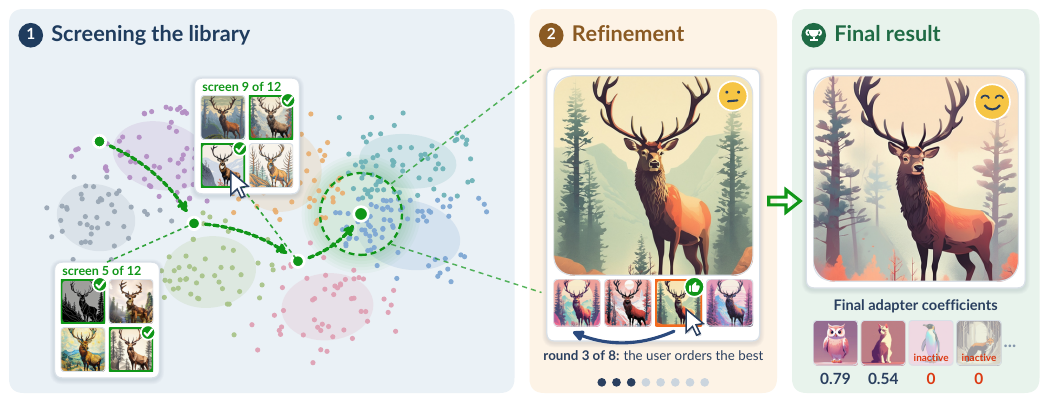}
  \caption{HimmBO lets a user explore merges of a whole adapter library ($N{=}1080$ here) through preference feedback.
  \emph{Screening} narrows the library to a small set of promising adapters using only precomputed images, so it requires no image generation during interaction (left).
  \emph{Refinement} then searches how to combine those adapters (middle) and returns the preferred merge and its coefficients (right).
}
  \label{fig:teaser}
}

\maketitle

\begin{abstract}
Fine-tuning has become a common way to customize text-to-image models, leading to community collections of over 100K lightweight adapters.
Adapters sharing a base model can be merged with continuous coefficients.
This turns a large collection into a high-dimensional continuous design space, which recent work explores interactively using preferential Bayesian optimization (PBO).
However, existing approaches are limited to a few dozen preselected adapters and require expensive image generation in every interaction round.
We introduce HimmBO, a two-stage method that scales preference-driven adapter merging to much larger collections while reducing image-generation cost.
Its screening stage recovers the active adapter set through level-set estimation over pre-generated images, requiring no image generation.
The refinement stage then optimizes merge coefficients in the recovered low-dimensional subspace.
Both stages use a style-aware kernel, allowing preference information to transfer between visually similar adapters.
In simulated-user image-matching experiments, HimmBO achieves higher final similarity than all comparisons while reaching a given similarity with under one tenth of the image generations required by prior work.
Its performance remains stable as the collection grows to 1080 adapters, while competing methods degrade substantially.

\end{abstract}

\input{sections/00_introduction}

\input{sections/01_related_work}

\input{sections/02_background}
\input{sections/03_method}

\input{sections/04_results}

\input{sections/05_conclusion}

\bibliographystyle{eg-alpha-doi}
\bibliography{references}

\input{sections/06_supplemental}

\end{document}

%% file: macros.tex
\usepackage{url}
\usepackage{xcolor}
\usepackage{booktabs}
\usepackage{wrapfig}
\usepackage{needspace}

\usepackage{amsmath}
\ifdefined\mathbb\else
  
  \usepackage{amssymb}
\fi
\usepackage{bm}

\DeclareMathOperator*{\med}{median}
\DeclareMathOperator{\Quant}{Quantile}
\DeclareMathOperator{\PCA}{PCA}
\DeclareMathOperator{\nrm}{normalize}

\newcommand{\x}{\mathbf{x}}
\newcommand{\z}{\mathbf{z}}

\newcommand{\Aset}{\mathcal{S}}                 %
\newcommand{\Dsp}{\mathcal{A}}                 %
\newcommand{\Dcmp}{\mathcal{D}}                 %
\newcommand{\dom}{\Delta_{B}}             %
\newcommand{\zhat}{\hat{\z}}
\newcommand{\Kdiv}{K^{\mathrm{div}}}

\newcommand{\ldiv}{\ell_{\mathrm{div}}}

\newcommand{\R}{\mathbb{R}}
\newcommand{\E}{\mathbb{E}}
\newcommand{\norm}[1]{\lVert #1 \rVert}

\newcommand{\W}{\mathbf{W}}
\newcommand{\balpha}{\bm{\alpha}}

\newif\ifsubmission \submissionfalse

%% file: sections/00_introduction.tex
\section{Introduction}
\label{sec:intro}

\begin{figure}[t]
  \centering
  \includegraphics[width=\linewidth]{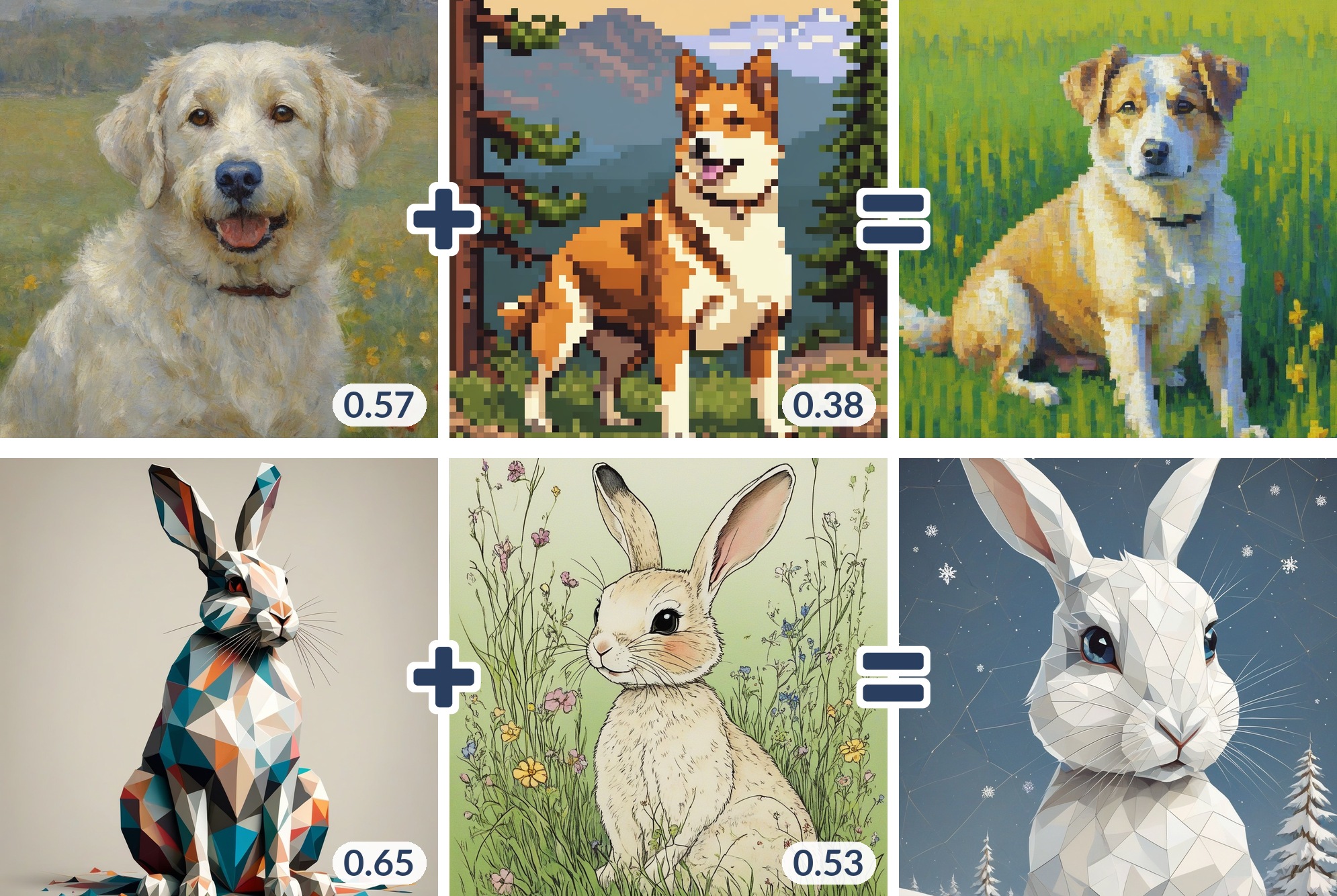}
  \caption{
  Both the selected adapters and coefficients determine the appearance of the merged image.
Each row shows two adapters used separately (left, middle) and a merge of the two (right).
}
  \label{fig:designspace}
\end{figure}

Fine-tuning has become a common way to customize text-to-image diffusion models for particular subjects, appearances, and artistic styles \cite{hu2022lora, ruiz2023dreambooth, zheng2024stylebreeder}.
The lightweight nature of parameter-efficient adaptations such as LoRA has also encouraged online communities to build and share large collections of fine-tuned models, which we refer to broadly as \emph{adapters}.
Beyond using adapters individually, users can also perform operations over multiple adapters to create new models and visual effects.
One such operation, and the focus of this work, is \emph{adapter merging}, where multiple adapters are combined with user-specified coefficients to produce a hybrid model that reflects their respective contributions \cite{ilharco2023editing}.
This turns a collection of $N$ adapters into an $N$-dimensional continuous design space (Fig.~\ref{fig:designspace}).
Even interfaces exposing only a handful of coefficients become difficult to operate as the number of parameters grows, with prior studies reporting substantial degradation beyond roughly 5--10 sliders \cite{dang2022ganslider}.
As community adapter collections continue to grow, with over 100K adapters on Civitai alone~\cite{luo2024stylus,sonmezer2025loraverse}, directly exploring the resulting space becomes increasingly impractical.

\begin{wrapfigure}{r}{0.32\linewidth}
  \centering
  \includegraphics[width=\linewidth,trim={2.5em 1em 0 1.5em}]{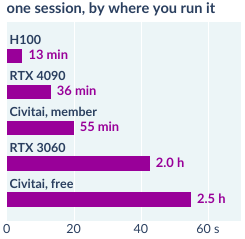}
\end{wrapfigure}
Selecting such coefficients can be viewed as optimizing a black-box user-preference function over the adapter-merging space.
GimmBO \cite{liu2026gimmbo} recently showed that preferential Bayesian optimization (BO) can solve this problem interactively for roughly 20--30 candidate adapters within 20 interaction rounds.
However, two practical limitations remain: each iteration requires computationally expensive image generation, and the candidate set must already be reduced to only a few dozen adapters.
Unlike conventional BO applications with near-instantaneous evaluations~\cite{koyama2017sequential,koyama2020sequential}, image generation makes even 20 rounds costly in wall-clock time, from minutes to hours depending on hardware or service (inset).
Obtaining such a small candidate set requires either manual curation or an external retrieval system \cite{luo2024stylus, sonmezer2025loraverse}, adding user effort in the former case and requiring a sufficiently accurate language description in the latter.
Our goal is thus to make preference-driven exploration practical over much larger adapter collections while reducing the image-generation cost.

We introduce HimmBO, a preference-driven optimization method that scales well with candidate adapter set size while remaining image-generation efficient.
HimmBO builds on GimmBO's two-stage strategy, which runs BO over all adapter coefficients inside a capped simplex that keeps merges sparse, reads the active set off the best merge found so far, then restarts BO in the lower-dimensional space those adapters define \cite{liu2026gimmbo}.
At higher dimensions, this strategy becomes error-prone because the second refinement stage relies on the sparsity pattern of a single tentative optimum.
Our key insight is that screening need not identify the optimum itself, but only reliably recover the active adapter set (Sec.~\ref{sec:screening}).
We explore several alternative Bayesian formulations for screening, including \emph{active learning} (AL) and \emph{level-set estimation} (LSE), which consider a broader portion of the preference landscape rather than concentrating on a single optimum.
We ultimately select LSE, which most accurately recovers the active adapters by drawing on multiple promising samples (Sec.~\ref{sec:eval-ablation}).
Crucially, LSE also admits a pool-based setting in which candidates are selected from a fixed \emph{sample library} rather than generated dynamically during optimization.
This allows us to pre-generate the corresponding images as a \emph{gallery}, eliminating image-generation wait time during screening (Sec.~\ref{sec:init}).
Finally, different adapters can produce similar visual results, so preference for one can inform preference for another similar adapter.
We capture this relationship with a \emph{style kernel} to further improve performance (Sec.~\ref{sec:surrogate}).

We evaluate HimmBO through quantitative simulated-user experiments on image-matching tasks, following GimmBO's evaluation protocol \cite{liu2026gimmbo} (Sec.~\ref{sec:results}).
In addition to final image similarity and ground-truth active-adapter recovery, we evaluate scalability and image-generation cost.
To test scalability, we increase the adapter collection size from $N=40$ to $120$, $360$, $720$, and $1080$.
HimmBO degrades gracefully with library size, reaching $0.889$ similarity at $N=1080$, only $0.018$ below its $0.907$ at $N=40$.
In contrast, the best alternative drops sharply from $0.903$ at $N=40$ to $0.828$ at $N=1080$, and GimmBO itself to $0.762$.
HimmBO is also substantially more generation-efficient, reaching $0.80$ similarity with $6.8\%$ of the image generations GimmBO needs to reach the same level.
Together, these results extend preference-driven model merging from a small, pre-curated adapter set to collections approaching the scale of contemporary community model libraries.

%% file: sections/01_related_work.tex
\section{Related Work}
\label{sec:related}

\paragraph{Adapter libraries and merging.}
Diffusion models generate images by gradually denoising random noise~\cite{ho2020ddpm,rombach2022ldm}, with large open models such as SDXL~\cite{podell2024sdxl} serving as bases for much of the surrounding community ecosystem.
Adapters are weight updates trained on top of such a base to add a subject or style~\cite{ruiz2023dreambooth,zhang2025survey,zheng2024stylebreeder}, including popular parameter-efficient forms such as LoRA~\cite{hu2022lora}.
Because they can be trained from only a handful of images and are cheap to store and share, community collections now contain over 100K adapters~\cite{luo2024stylus,sonmezer2025loraverse}.
Adapters from the same base can be combined by adding their weight updates, following a long history of weight averaging across related models~\cite{izmailov2018swa,wortsman2022modelsoups}.
This works because fine-tuned models often remain in the same low-loss region, allowing their updates to be added and scaled~\cite{frankle2020linear,ilharco2023editing}.
When updates interfere, more elaborate rules can reweight or sparsify them~\cite{matena2021fisher,yadav2023ties,yu2024dare}, but simple linear merging is what community tools commonly implement and what we use as well (Sec.~\ref{sec:background}).
In image generation, merging is usually applied to a small, fixed set of adapters to compose several concepts or combine a subject with a style~\cite{gu2023mixofshow,po2024orthogonal,dalva2025lorashop,shah2024ziplora,frenkel2024blora,ouyang2025klora}.
These methods assume the participating adapters are given in advance, with coefficients either fixed by the method or fitted to a predefined target.
GimmBO instead treats the coefficients as a search problem, using user preference comparisons in place of an explicit target~\cite{liu2026gimmbo}.
Its candidate set, however, is still small and chosen beforehand.
Our work extends this interactive search from a small preselected set to an entire adapter library.

\begin{figure}[t]
  \centering
  \includegraphics[width=\linewidth]{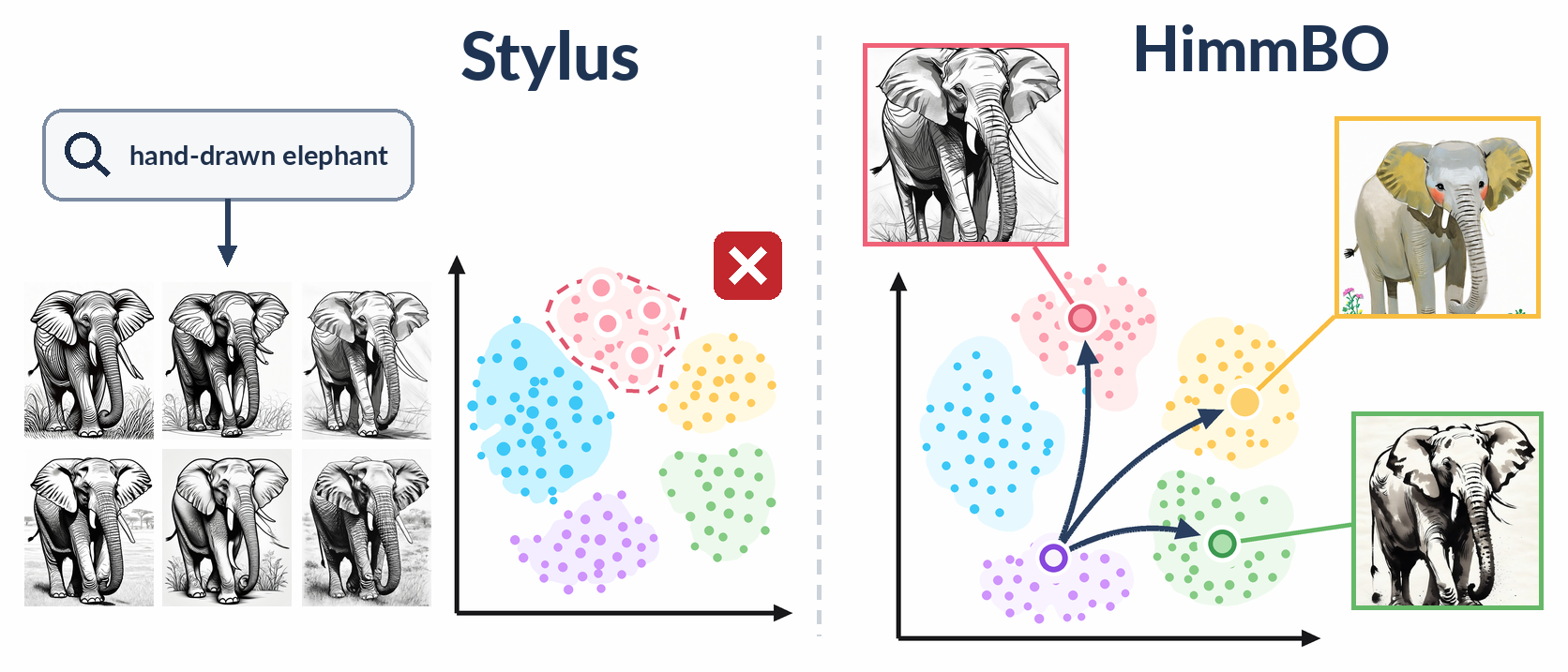}
  \caption{
  Retrieval depends on a query specified before search begins.
Here, “hand-drawn” confines Stylus~\cite{luo2024stylus} to a narrow sketch-like cluster, while HimmBO lets the user interactively explore broader related styles.
}
  \label{fig:styluscompare}
\end{figure}

\paragraph{Selecting adapters from a large collection.}
When a collection contains hundreds or thousands of adapters, a common first step is to retrieve a small subset relevant to the user's query.
Stylus retrieves adapters from a text query~\cite{luo2024stylus}, while LoRAverse additionally promotes coverage across the collection~\cite{sonmezer2025loraverse}.
Other work retrieves directly from model weights, either using them as style descriptors~\cite{liu2024loradescriptor} or aligning them with text for prompt-based retrieval~\cite{li2025autolora}.
Similar ideas apply to language adapters, using a task loss or a few examples instead of a prompt~\cite{huang2024lorahub,zhao2024loraretriever,chronopoulou2023adaptersoup}.
These methods choose the adapter set once from a query specified before search begins, which may be difficult for users who cannot yet state their preference explicitly.
We therefore make adapter selection part of the interactive loop and compare against retrieval for the same task (Sec.~\ref{sec:results}).

\paragraph{Human-in-the-loop design optimization.}
When quality is subjective, users must remain in the loop, and prior work has explored interfaces such as sliders and galleries.
Direct slider control becomes difficult beyond roughly five to ten parameters~\cite{dang2022ganslider}, while systems that learn subjective quality from human judgments still operate below the dimensionality we target~\cite{koyama2014crowdpowered,desai2019geppetto,shimizu2020designadjectives}.
Galleries instead present outputs rather than parameters, either arranging samples from the design space for browsing~\cite{marks1997designgalleries,ngan2006brdf,shapira2009appearance} or selecting from a fixed image database based on the user's choices~\cite{cox2000pichunter}.
They can remain responsive by precomputing expensive results and reusing them during interaction~\cite{shugrina2015fabforms,schulz2017cad}.

BO provides a sample-efficient backend for human-in-the-loop design~\cite{brochu2010interactive,koyama2017sequential,koyama2020sequential} and has been applied to domains such as melodies, fonts, and game levels~\cite{zhou2020melody,kadner2021adaptifont,tatsukawa2025fontcraft,khajah2016games}, alongside work on how users and optimization systems cooperate~\cite{koyama2022assistant,mo2024cooperative,niwa2025cooperative}.
These methods typically propose new parameter settings each round, so their outputs must be generated during interaction rather than fully precomputed.
GimmBO falls into this category~\cite{liu2026gimmbo}, and our refinement stage similarly returns to continuous BO once the candidate adapters have been reduced.
However, screening has a different goal: rather than finding a single optimum, it must identify a set of promising adapters across the preference landscape.
We use pool-based level-set estimation (LSE) for screening to achieve better active-adapter recovery while allowing all screening images to be precomputed (Sec.~\ref{sec:init}\&\ref{sec:eval-ablation}).

\paragraph{Goal-driven learning.}
The human-in-the-loop methods above are examples of goal-driven learning, where each query is chosen from the current surrogate to serve a particular objective~\cite{di2024active,polke2026adaptive}.
BO seeks a single best point~\cite{srinivas2010gpucb}, AL an accurate surrogate through uncertainty- or information-based queries~\cite{mackay1992information,seo2000active,houlsby2011bald,binois2019imspe}, and LSE the set of candidates above a level~\cite{bryan2005active,gotovos2013active,bogunovic2016truvar}.
LSE therefore best suits our screening stage, which recovers a set of promising adapters rather than a single optimum (Sec.~\ref{sec:eval-ablation}).
With preferential rather than scalar feedback, latent utility can be modeled with a probit likelihood~\cite{chu2005preference}, with preferential Bayesian optimization defining acquisition rules on this model~\cite{gonzalez2017pbo,astudillo2023qeubo}.
For batched queries, diversity may be implicit or explicitly encouraged, for example with determinantal point processes~\cite{kulesza2012dpp,kathuria2016dpp}.
Large adapter libraries also introduce high dimensionality, addressed by prior work through linear embeddings~\cite{wang2016rembo,raponi2020pcabo,chiu2020subspace}, trust regions~\cite{diouane2023trego}, or sparse coordinate selection~\cite{eriksson2021saas,liu2023sparsebo}.
These strategies can ease search in a high-dimensional coefficient space, but its dimension still grows linearly with the number of adapters.
We instead use a fixed-dimensional style space where nearby adapters share feedback (Sec.~\ref{sec:background}).

%% file: sections/02_background.tex
\section{From a curated set to a library}
\label{sec:background}

Adapters fine-tuned from the same base model can be merged.
Given a library of $N$ adapters and a coefficient vector $\balpha\in\R^{N}$ that assigns each adapter a strength, a merging procedure combines the adapters' weight updates into a single model, and that model generates images that mix the adapters' visual effects.
Throughout the paper we use the standard linear rule
\begin{equation}
  \W(\balpha)=\W_0+\textstyle\sum_{i}\alpha_i\Delta\W_i,
  \label{eq:merge}
\end{equation}
where $\W_0$ is the base model and $\Delta\W_i$ the $i$th adapter's weight update~\cite{hu2022lora,liu2026gimmbo}.
It is simple, effective, and what community tools implement.
However, our method does not depend on this particular merging rule and can be applied seamlessly to alternatives.

GimmBO~\cite{liu2026gimmbo} turns the choice of $\balpha$ into an interactive loop: each round it generates a batch of candidate merges, the user ranks the images, the ranking updates a surrogate function of the user's preference over $\balpha$, and an acquisition function picks the next batch.
This works well when the candidate set is small and curated, typically 20--30 adapters chosen beforehand.
Public collections, however, contain thousands of adapters, so users must solve a more fundamental problem before tuning coefficients: deciding which adapters to use in the first place.
We build on two observations to extend the same interactive loop to answer both questions.

\needspace{12\baselineskip}%
{\setlength{\columnsep}{6pt}\setlength{\intextsep}{4pt}%
\paragraph{Preference has several peaks.}
\begin{wrapfigure}{r}{0.48\linewidth}
  \centering
  \includegraphics[width=\linewidth]{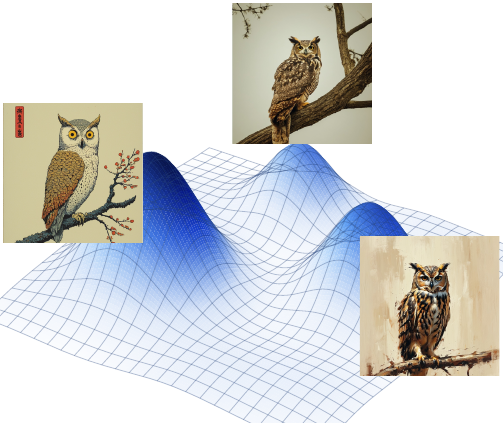}
\end{wrapfigure}
A user who merges style adapters rarely has a single look in mind: the same user may find a woodblock print, a photograph, and an oil painting all close to what they want (inset), producing several distinct peaks of preference across the library.
A search for a single best point can lock onto whichever peak is favored by its first few comparisons and keep querying there, especially in the high-dimensional setting we target, where only a small fraction of the space can be explored.
Such a miss is silent---a user who never sees a style cannot ask for more of it---and can exclude adapters needed for a final merge spanning several peaks.
Unlike GimmBO, whose first screening stage relies on a single tentative optimum and often captures only one peak, our search seeks broader coverage by identifying a set of promising points spanning multiple peaks (Sec.~\ref{sec:method}).
\par}

\begin{figure}[t]
  \centering
  \includegraphics[width=0.9\linewidth]{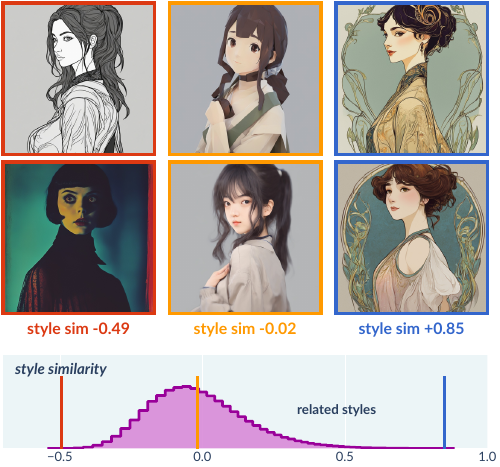}
  \caption{
  Adapter identities do not reflect visual distance: independently trained adapters can be unrelated, related, or nearly identical in appearance.
Example pairs are shown at their measured style similarities (top), with the full pairwise distribution across the library below.
}
  \label{fig:stylecoef}
\end{figure}

\paragraph{Preference depends on style, not coefficients.}
Users judge generated images by appearance, and visually similar styles occur throughout large adapter libraries.
Independently trained adapters can produce nearly indistinguishable looks (Fig.~\ref{fig:stylecoef}, top).
Yet coefficient space does not capture this structure because each adapter occupies its own axis, making all pairs equally distant regardless of visual similarity and preventing a preference model over $\balpha$ from transferring feedback to lookalikes.
The feedback required therefore grows with the number of adapters.
We instead model preference in a style space measured from generated images, allowing feedback on one adapter to inform others with similar styles.

%% file: sections/03_method.tex
\section{Method}
\label{sec:method}

Given a library of $N$ adapters and a merging rule, our goal is to find the merge that best matches the user's preference:
\begin{equation}
  \max_{\balpha\in\Dsp}\; f\big(I(\balpha)\big),
  \label{eq:problem}
\end{equation}
where $\Dsp=\R^{N}$ is the $N$-dimensional design space, $I(\balpha)$ is the image generated by the merged model, and $f$ is the user's \emph{preference function}, assigning each image a scalar utility reflected in the user's choices.
The adapters with nonzero coefficients form the merge's \emph{active set}, $\Aset=\operatorname{supp}(\balpha)$.
In practice, merges tend to have small active sets~\cite{liu2026gimmbo}, so we follow this observation and retain $m$ adapters for the refinement stage.
GimmBO~\cite{liu2026gimmbo} extracts the active set from the support of the best merge found after several PBO iterations (10 by default~\cite{liu2026gimmbo}).
At library scale, however, relying on a single tentative optimum becomes unreliable: it can miss adapters from other promising regions, making active-set recovery the central challenge.

Our answer is a single preference posterior driving two stages, shaped by the two observations in Sec.~\ref{sec:background}.
The \emph{screening stage} searches the full library for a small active set covering styles the user may still prefer, rather than committing to a single tentative optimum.
The \emph{refinement stage} then searches merges within that set.
Two design choices make this practical at library scale.
A \emph{gallery} generated once per library contains every image shown during screening, eliminating image generation during screening (Sec.~\ref{sec:init}).
A style kernel lets feedback on one adapter inform visually similar adapters, so each batch of $q$ images informs the full library (Sec.~\ref{sec:surrogate}).
Both stages share and update the same posterior, so all feedback carries from screening into refinement.

\subsection{Style representation}
\label{sec:style}

\begin{figure*}[t]
  \centering
  \includegraphics[width=\linewidth]{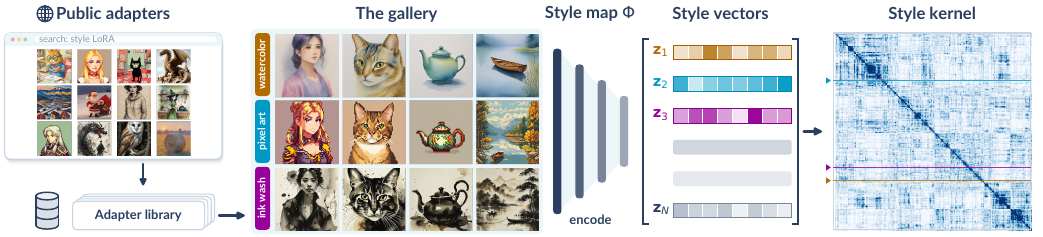}
  \caption{
  Each adapter in the public library is generated alone on the same four prompts to form the gallery.
The style map encodes these images into a style vector, and the resulting style kernel captures visual similarity across the library.
  }
  \label{fig:stylemap}
\end{figure*}

Users judge generated images by their appearance, not by the coefficients that produce them, so we model the user's preference as $f(\z)$, where $\z$ represents visual style rather than the merging coefficients directly.
We obtain $\z$ through a \emph{style map} $\Phi$ applied to the generated image (Fig.~\ref{fig:stylemap}):
\begin{equation}
  \z=\Phi\big(I(\balpha)\big)\in\R^{d}.
  \label{eq:premise}
\end{equation}
Thus, the coefficients affect preference only through the appearance of the image they produce.
We construct $\Phi$ from perceptual image features by centering features within each prompt to suppress shared content and composition, then normalizing and projecting with PCA (details in App.~\ref{sec:supp-stylemap}).
The perceptual encoder is interchangeable, and we use DreamSim~\cite{fu2023dreamsim} by default (alternative encoders are compared in App.~\ref{sec:supp-encoder}).
Its output dimension is fixed with respect to library size, and PCA further compresses it using a fixed variance threshold, yielding a small dimension $d$ that changes little as $N$ grows.
During screening, gallery images are pre-generated, so $\z$ is measured directly with $\Phi$.
During refinement, $\z$ is predicted from $\balpha$ for acquisition and then measured directly once the proposed merge is generated (Sec.~\ref{sec:refine}).

\subsection{Preferential surrogate}
\label{sec:surrogate}

We estimate the user's preference with a latent Gaussian-process surrogate over an input representation $\x$,
$\hat{f}\sim\mathcal{GP}\big(0,k(\x,\x')\big)$, whose posterior given the comparisons $\Dcmp$ below has mean $\mu$ and standard deviation $\sigma$.
Because users provide choices rather than numeric scores, all feedback is converted into pairwise comparisons and accumulated in a single data set $\Dcmp$.
During screening, the user simply \emph{marks} the preferred images per batch, producing comparisons between marked and unmarked images.
During refinement, where only a small candidate set remains, the user instead \emph{ranks} their top choices, which similarly yields pairwise comparisons.
We infer latent utilities from these comparisons through a standard probit likelihood and approximate the posterior with Laplace approximation~\cite{chu2005preference}.
Kernel hyperparameters are re-fit each round by maximum a posteriori estimation.
See Sec.~\ref{sec:impl} for details.

What should $\x$ represent?
The conventional choice, also used by GimmBO, is the coefficient vector itself, $\x=\balpha$.
This works for a small candidate set but becomes a bottleneck at library scale.
Each adapter occupies its own axis, so feedback on one adapter cannot inform visually similar others, and the amount of user feedback needed grows rapidly with $N$.
We instead represent each observed candidate by both its style feature and merging coefficients,
$\x=[\,\z\Vert\balpha\,]$, and use the additive kernel
\begin{equation}
  k(\x,\x')=\sigma_s^{2}\,e^{-\norm{\z-\z'}^{2}/2\ell_s^{2}}
         +\sigma_I^{2}\,e^{-\norm{\balpha-\balpha'}^{2}/2\ell_I^{2}},
  \label{eq:kernel}
\end{equation}
with channel variances $\sigma_s^{2},\sigma_I^{2}$ and length scales $\ell_s,\ell_I$.
The style channel transfers feedback between visually similar adapters, allowing each batch to inform the model across the library.
The coefficient channel preserves distinctions between different merges that produce similar styles.
We use a sum so that similarity in either channel can transfer preference information.

\needspace{13\baselineskip}%
{\setlength{\columnsep}{6pt}\setlength{\intextsep}{4pt}%
\begin{wrapfigure}{r}{0.38\linewidth}
  \centering
  \includegraphics[width=\linewidth, trim={0.2em 0.5em 0 0.3em}]{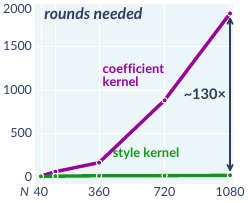}
\end{wrapfigure}
We quantify the effect of the style kernel in a simulated-user experiment, measuring how many feedback rounds are needed before we recover half of the active adapters in the ground-truth merge.
With a coefficient-only kernel, this feedback cost grows rapidly with library size.
With the style kernel, it remains nearly flat and is about $130\times$ lower at $N{=}1080$ (experiment details in App.~\ref{sec:supp-reach}).
\par}

\subsection{Two-stage optimization}
\label{sec:hier}

\mbox{}\par\vspace{-\baselineskip}%
{\setlength{\columnsep}{6pt}\setlength{\intextsep}{4pt}%
\begin{wrapfigure}{r}{0.45\linewidth}
  \centering
  \includegraphics[width=\linewidth,trim={0 0.5em 0 0.5em}]{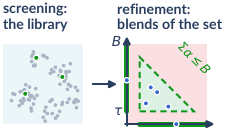}
\end{wrapfigure}
\noindent
HimmBO separates the search into two stages with distinct goals (inset).
Screening searches the full library to identify an active set of $m$ adapters from the $N$ candidates, while refinement continuously optimizes the merge coefficients within that set.
The active set produced by screening defines the reduced search space for refinement.
Both stages share and update the same posterior (Sec.~\ref{sec:surrogate}), so feedback collected during screening carries into refinement.
\par}

\subsubsection{Precomputed gallery}
\label{sec:init}

Given a coefficient vector $\balpha$ for an adapter library, we define its \emph{gallery} as a set of images generated by the corresponding merged model (Fig.~\ref{fig:stylemap}).
As described in Sec.~\ref{sec:style}, these images also serve as inputs to the style map $\Phi$, yielding the corresponding style feature $\z$.
Before any user interaction, we precompute galleries for a set of coefficient samples using the same fixed prompt set (details in App.~\ref{sec:supp-gallery}), associating each $\balpha$ with its gallery and resulting style feature $\z$.
Each such sample therefore provides the full surrogate input $\x=[\,\z\Vert\balpha\,]$ (Sec.~\ref{sec:surrogate}).
If subsequent interaction selects only among these samples, their gallery images can be shown directly, requiring no image generation during interaction.

\begin{figure}[t]
  \centering
  \includegraphics[width=\linewidth]{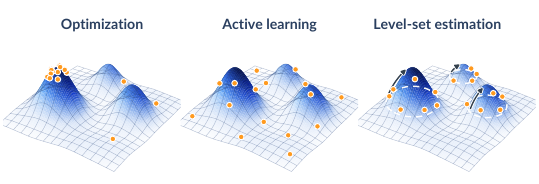}
  \caption{
  With the same preference landscape (higher is more preferred) and query budget (dots), BO focuses on one peak, AL spreads queries broadly, and LSE targets uncertain boundaries around multiple promising peaks.
Boundary evidence can establish a peak as above the level, retaining unqueried adapters (arrows).
}
  \label{fig:acquisition}
\end{figure}

\subsubsection{Screening}
\label{sec:screening}

The goal of screening is to deliver an active set of $m$ adapters for refinement.
Because refinement cannot recover adapters omitted from this set, screening must cover promising regions rather than identify a single optimum.
We thus consider active learning (AL) and level-set estimation (LSE), which target broader regions than a single maximizer.
Although developed separately from BO, all three can be viewed as goal-driven learning over a shared surrogate with different acquisition goals~\cite{di2024active,polke2026adaptive,gotovos2013active} (Fig.~\ref{fig:acquisition}).

The three formulations differ in what they seek from the same preference posterior.
Bayesian optimization (BO) seeks a maximizer,
$\x^\star=\arg\max_{\x\in\mathcal X} f(\x)$.
AL instead seeks an accurate surrogate over the domain, commonly framed as reducing integrated posterior variance,
$\int_{\mathcal X}\sigma^2(\x)\,d\x$~\cite{di2024active,polke2026adaptive}.
LSE seeks the set of candidates whose utility exceeds a level~\cite{gotovos2013active},
\begin{equation}
  H^\star=\{\,i:\ f(\z_i)\ge h^\star\,\},
  \label{eq:levelset}
\end{equation}
where $h^\star$ is a quantile of the utilities rather than a fixed threshold, since preference provides no absolute utility scale.

Both AL and LSE admit a \emph{pool-based} setting, where queries are selected from a fixed candidate set.
We instantiate the pool with one unit-strength sample per adapter, with its gallery and style feature $\z_i$ precomputed as in Sec.~\ref{sec:init}.
This is particularly practical because adapter platforms already maintain representative galleries for individual adapters.
Our pool reuses images that would already be generated and stored, adding no extra image-generation or gallery-storage cost.
Each round can then score the full pool and show selected samples directly from their galleries, requiring no image generation during screening.

These objectives produce different query patterns (Fig.~\ref{fig:acquisition}).
BO can concentrate around an early peak, leaving others unexplored.
This is especially problematic in high dimensions, where a limited budget may miss peaks containing adapters needed by the final merge.
AL covers the landscape broadly but spends queries in low-utility regions the user has rejected.
LSE instead concentrates on uncertain level-set membership around multiple promising peaks, retaining candidates estimated above the level.
Both AL and LSE recover the active set more accurately than BO as the library grows; LSE performs best and is used in our method (Sec.~\ref{sec:eval-ablation}).

Screening proceeds in rounds: the current posterior scores the candidate pool, $q$ samples are shown from their precomputed galleries, and the user's selections update the posterior.
We initialize the first batch with one random candidate and choose the remaining $q-1$ by farthest-first traversal in style space for broad coverage.
Empirically, this initialization has little effect on the final result.

For subsequent rounds, we score each pool candidate by uncertainty in its level-set membership using the straddle rule~\cite{bryan2005active},
\begin{equation}
  a(i)=1.96\,\sigma_i-\lvert \mu_i-\hat h\rvert,
  \qquad
  \hat h=\Quant_{\theta}\big(\{\mu_j\}\big),
  \label{eq:straddle}
\end{equation}
where $i$ indexes the pool candidates and $\mu_i,\sigma_i$ are their posterior mean and standard deviation.
We re-estimate $\hat h$ each round so that a fraction $1-\theta$ of the candidates lies above it, while $1.96$ follows the original straddle rule and corresponds to a $95\%$ posterior interval.
Straddle is a special case of LSE~\cite{gotovos2013active} that favors candidates with uncertain level-set membership.
Following LSE, we partition candidates into those confidently above, confidently below, and still uncertain, and query from the uncertain set.
We use only the LSE components compatible with preferential feedback, leaving broader adaptation of LSE to this setting as future work (Sec.~\ref{sec:conclusion}).

Each round selects the $q$ highest-scoring candidates, subject to a mutual-exclusion radius in style space so that the batch does not spend multiple queries on lookalikes.
We set this radius to the style length scale $\ell_s$ of Eq.~\eqref{eq:kernel}, within which the posterior already treats two candidates as near-copies of one query.

When screening ends, we form the estimated set of candidates above the level,
$\hat H=\{\,i:\mu_i\ge\hat h\,\}$, and select $m$ adapters from it for refinement.
Because $\hat H$ can span multiple preference peaks, the $m$ slots are shared across them so that promising regions retain representation.
Through the style kernel, an adapter need not be shown directly to enter $\hat H$: comparisons on visually similar adapters can raise its posterior mean above the level.
Details are in App.~\ref{sec:supp-lse}.

\subsubsection{Refinement}
\label{sec:refine}

Refinement turns the screened active set into a final merge by optimizing how its $m$ adapters are combined.
We search over
\begin{equation}
  \dom=\big\{\balpha\in\R^{m}:\ \alpha_i\in\{0\}\cup[\tau,1],\
  \textstyle\sum_i\alpha_i\le B\big\},
  \label{eq:domain}
\end{equation}
a sparse, capped simplex following the practical constraints used by GimmBO~\cite{liu2026gimmbo}.
We set $\tau=0.1$ to exclude coefficients with negligible visual effect, while the cap $B$ limits the total merge strength to avoid degraded generations.

Refinement follows GimmBO's interaction protocol, with our own modifications.
Starting from the posterior learned during screening, each round we obtain $q$ merges from $\dom$, generate their images, ask the user to rank their top choices, and update the posterior.
Because refinement ultimately seeks a single best merge, we return to the BO formulation and optimize
\begin{align}
  \mathop{\text{max}}_{\{\balpha_1,\dots,\balpha_q\}} & \quad \E_{\phi}\big[\max_{j=1,\dots,q}\phi(\x_j(\balpha_j))\big]
  +\gamma\log\det\big(\Kdiv+\epsilon I\big),
  \label{eq:acq} \\
  \text{subject to:}&\quad \balpha_j \in \dom \text{ for } j=1, \dots, q,
\end{align}
where the first term is the standard expected-maximum batch acquisition~\cite{wilson2017reparam}, with $\phi$ chosen as UCB,
$\phi(\x)=\mu(\x)+\lambda\sigma(\x)$~\cite{srinivas2010gpucb}, and the second is a diversity term with weight $\gamma = 0.02$ and $\epsilon=10^{-3}$ in all experiments.
Because a proposed merge has no image yet, we approximate its style feature from the gallery.
Motivated by the approximately additive visual effects of LoRA merging, we predict the style feature as $\zhat(\balpha)=\sum_{i\in\Aset}\alpha_i\,\z_i$, summing the active adapters' style features weighted by their coefficients (App.~\ref{sec:supp-acqopt}).
This gives the surrogate input
$\x=[\,\zhat(\balpha)\Vert\balpha\,]$
used for acquisition.
Once the merge is generated, its measured style feature $\z$ replaces $\zhat(\balpha)$ in the posterior update.
We construct the diversity term following determinantal point processes~\cite{kulesza2012dpp}, with
$\Kdiv_{jk}=e^{-\norm{\bar\z_j-\bar\z_k}^{2}/2\ldiv^{2}}$
measuring similarity between normalized predicted style features
$\bar\z_j=\zhat(\balpha_j)/\norm{\zhat(\balpha_j)}$.
The standard batch term already encourages separation through posterior covariance, but can still yield visually redundant candidates.
We add this log-determinant term to spread each batch across style rather than coefficient axes, increasing diversity while leaving the session's final similarity essentially unchanged (Fig.~\ref{fig:mechanism}, right; Sec.~\ref{sec:eval-ablation}).

\needspace{10\baselineskip}%
{\setlength{\columnsep}{6pt}\setlength{\intextsep}{4pt}%
\begin{wrapfigure}{r}{0.58\linewidth}
  \centering
  \includegraphics[width=\linewidth]{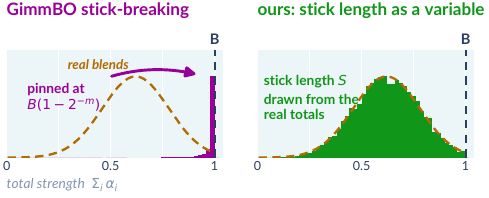}
\end{wrapfigure}
Finally, we modify GimmBO's stick-breaking parameterization to remove its strong bias toward the cap $B$.
GimmBO parameterizes $\dom$ with variables in the unit cube so that L-BFGS-B can be applied directly.
Conceptually, it starts with a stick of length $B$, lets each of the $m$ adapter coordinates take a fraction of the remaining stick, and drops the final remainder as a slack coordinate.
With uniform cube samples, the expected remainder shrinks to $2^{-m}B$, so the retained coefficients have total
$\E[\sum_i\alpha_i]=(1-2^{-m})B$.
At $m=8$, this is within $1\%$ of $B$, concentrating samples near the cap even though real community merges span substantially lower totals (inset, left).
We instead make the stick length $S\le B$ itself a variable.
During sampling, $S$ is drawn from the empirical distribution of coefficient totals in public merges; during optimization, it becomes an $(m{+}1)$-th decision variable, allowing the total budget to be searched directly (inset, right).
The floor $\tau$ of Eq.~\eqref{eq:domain} is enforced within the same map.
We use the same public usage statistics to define this distribution and to construct our test cases, but the two uses are independent (see App.~\ref{sec:supp-acqopt}).
\par}

\subsection{Implementation details}
\label{sec:impl}

We implement HimmBO in Python with BoTorch~\cite{balandat2020botorch} and GPyTorch~\cite{gardner2018gpytorch}.
Screening returns $m=8$ active adapters for refinement.
We re-fit the kernel hyperparameters of Eq.~\eqref{eq:kernel} each round by maximum a posteriori estimation under log-normal priors centered on offline-calibrated values, and use a smaller exploration weight than GimmBO in refinement.
The remaining constants, readout details, and timings are given in App.~\ref{sec:supp-impl}.

As in GimmBO, all images during refinement are generated from a shared anchor image using SDEdit~\cite{meng2022sdedit}, keeping content fixed so the user primarily compares style.
In practice, optimization overhead is small relative to image generation.
A screening round, including posterior fitting and scoring the full candidate pool, takes about $0.3$\,s, while refinement requires about $1$\,s for posterior updating and acquisition optimization.
Thus, search computation adds negligible interaction delay.

%% file: sections/04_results.tex
\section{Evaluations}
\label{sec:results}

We evaluate with simulated users on a standard matching task (Sec.~\ref{sec:eval-quant}) and demonstrate three extensions (Sec.~\ref{sec:eval-ext}).

\subsection{Quantitative experiments}
\label{sec:eval-quant}

\begin{figure*}[t]
  \centering
  \includegraphics[width=\linewidth]{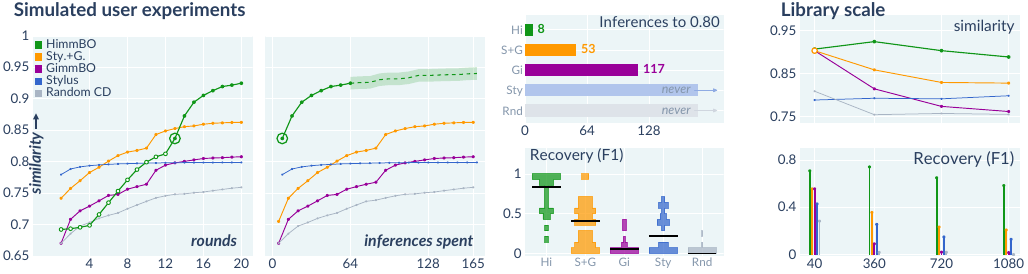}
  \caption{
  Across 24 targets and five seeds, HimmBO improves through screening and refinement when measured by either interaction rounds or inferences at $N{=}360$ (left).
Hollow screening points are diagnostic readouts excluded from the inference budget, and the ring marks the start of refinement.
The two smaller panels report the number of inferences needed to reach $0.80$ similarity and the accuracy of target-adapter recovery (F1).
Across nested libraries, HimmBO maintains high similarity as $N$ grows while most accurately recovering the target adapters at every scale (right).
}
  \label{fig:e2e}
\end{figure*}

\paragraph{Task and protocol.}
We evaluate HimmBO with a matching task, a standard protocol for human-in-the-loop optimization~\cite{brochu2010interactive,koyama2017sequential,koyama2020sequential,liu2026gimmbo}.
We construct 24 test target images with different prompts, each generated from a hidden merge of multiple library adapters with strengths $\sum_i\alpha_i$ drawn from the empirical distribution observed in community merges.
For each ground-truth (GT) target, the simulated user judges the similarity of candidate images generated with the same prompt to the GT and provides only preference feedback to the method.
We evaluate on each target over five random seeds.
The simulated user evaluates each shown image by its similarity to $I_{\mathrm{GT}}$, using DreamSim~\cite{fu2023dreamsim} as the preference function, $f(I(\balpha),I_{\mathrm{GT}})=\mathrm{DreamSim}(I(\balpha),I_{\mathrm{GT}})$, following GimmBO's simulated-user protocol~\cite{liu2026gimmbo}.
During screening, we run $T_1{=}12$ rounds with $q{=}8$ precomputed gallery images per round, and the user selects the $c{=}4$ most similar.
Because screening operates entirely on precomputed galleries, these 12 rounds collect preference feedback without any new image generation.
During refinement, we run $T_2{=}8$ rounds, each generating $q{=}8$ merges and showing them alongside the previous round's best, from which the user ranks the top five.
We set these budgets taking user-facing cost into account, rather than performance saturation alone: screening mainly costs interaction effort, whereas refinement also incurs waiting time for image generation.
This gives 64 inferences per session, versus 165 in GimmBO's published configuration.
Because image generation dominates latency and all other computation takes only seconds on the CPU, the lower inference budget directly reduces the user's waiting time.
A session then takes about $15$ minutes versus $37$ for GimmBO on a single GPU.

\paragraph{Library.}
We curate \emph{SDXL-Style-1K}, a library of 1{,}080 SDXL LoRA style adapters.
We crawl public SDXL adapters on Civitai, filter them by tags and usage statistics, use a language model to verify that each changes style rather than content, and manually inspect standardized generations (App.~\ref{sec:supp-library}).
We generate each adapter's gallery once before interaction, using four fixed subjects with one image per subject.
For scalability experiments, we form nested libraries of $N{=}40$, 120, 360, 720, and 1080 adapters, each containing all smaller libraries.
Our main performance analysis uses the $N{=}360$ library, while Sec.~\ref{sec:eval-scale} evaluates performance across library sizes.
We will release the library manifest, curation pipeline, and test inputs.

\paragraph{Comparisons.}
We compare HimmBO against GimmBO, retrieval-based alternatives, and a slider-style baseline.
We evaluate GimmBO directly on the test library using its published configuration.
For retrieval, we use Stylus~\cite{luo2024stylus}, which retrieves adapters from text.
To adapt it to our image-matching task, we caption the target GT with a vision-language model and use that caption as the query, giving Stylus additional GT information unavailable to other methods.
At each round, we rerun Stylus with a new random seed to produce a new image batch, from which the user keeps the best.
We also include Stylus+GimmBO, following GimmBO's retrieval extension~\cite{liu2026gimmbo}: Stylus retrieves the top 40 adapters, then GimmBO runs its full method on this reduced set.
For the slider-style baseline, we include random coordinate descent, which adjusts one coefficient at a time over eight evenly spaced values.
We omit other slider-style baselines because they already performed poorly at $N{=}20$~\cite{liu2026gimmbo} and degrade further as $N$ grows.
GimmBO and random coordinate descent share the same five initial images and thus start from the same point; the remaining methods start from candidates produced by screening or retrieval.
Stylus and random coordinate descent evaluate eight candidates per round to match HimmBO's batch size, while GimmBO retains its published interaction protocol.

\paragraph{Metrics.}
\emph{Similarity} is measured by DreamSim~\cite{fu2023dreamsim} between the final generated image and the GT, following GimmBO~\cite{liu2026gimmbo}.
Scores of $0.9$ and $0.95$ correspond roughly to a match and a near-exact match, respectively.
\emph{Sparsity recovery} is measured by the F1 score between the adapters active in the final merge and the GT's true components~\cite{liu2026gimmbo}.
It penalizes both missed and redundant adapters, distinguishing whether a method recovers the true components or instead matches the GT using visually similar alternatives.
The style map uses DreamSim by default, but HimmBO still outperforms GimmBO when both it and the simulated user use other encoders, such as TPIPS~\cite{wang2026tpips} (App.~\ref{sec:supp-encoder}).

\subsubsection{Simulated-user comparison}
\label{sec:eval-main}

We set up the simulated user to see the hidden GT image and judge candidate images by DreamSim similarity, while the method receives only the resulting preference feedback, following GimmBO's matching protocol.
We analyze each session in two ways (Fig.~\ref{fig:e2e}, left).
First, we plot similarity against feedback rounds, measuring the user interaction budget.
To visualize screening progress, after each round we read out the current active set, generate one batch of $q{=}8$ merges using the refinement acquisition on the feedback so far, and plot the best result (hollow dots).
These diagnostic generations are excluded from HimmBO's inference budget.
Second, we plot the same progress against inferences, which captures the main source of user waiting time.
GimmBO evaluates efficiency in rounds, capturing interaction effort but not inference-induced waiting time.
HimmBO separates these costs: screening advances the round count using precomputed images without spending inferences, so the two axes differ.

\begin{figure*}[t]
  \centering
  \includegraphics[width=\linewidth]{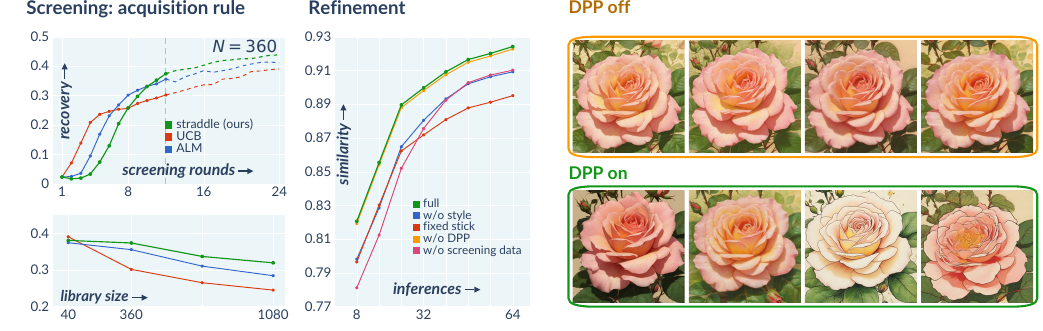}
  \caption{
  Ablation of HimmBO's screening and refinement components.
For screening, we compare UCB, ALM~\cite{mackay1992information,seo2000active}, and straddle~\cite{bryan2005active}; straddle overtakes the alternatives within twelve rounds at $N{=}360$ and leads from $N{=}360$ onward (left).
For refinement, all variants use the same active set: \emph{w/o style} removes the style kernel, \emph{fixed stick} fixes total strength to $B$, \emph{w/o DPP} removes diversity, and \emph{w/o screening data} discards screening comparisons (middle).
Fixed stick causes the largest similarity loss, while removing DPP mainly yields visually similar proposals (right).
}
  \label{fig:mechanism}
\end{figure*}

Against rounds, HimmBO improves steadily through screening, passes GimmBO before screening ends, and overtakes Stylus+GimmBO shortly after refinement begins, reaching the highest final similarity.
Stylus+GimmBO is the strongest alternative and benefits from GT-derived auto-captioning for retrieval.
Stylus alone starts strongly but quickly plateaus around $0.80$, while GimmBO improves more gradually and random coordinate descent remains lowest.
Against inferences, the ordering is similar, but HimmBO's advantage is clearer because screening requires no image generation.
HimmBO reaches $0.80$ after only 8 inferences, compared with 53 for Stylus+GimmBO and 117 for GimmBO; Stylus and random coordinate descent never reach this level.
A HimmBO session stops after 64 inferences, and the dashed continuation shows further improvement under the same inference budget as the comparisons.
Sparsity recovery further explains these differences.
HimmBO recovers the most GT adapters, followed by Stylus+GimmBO, whose subsequent search cannot recover components missed by Stylus.
Stylus alone recovers some GT components, whereas GimmBO recovers almost none despite moderate image similarity.
Instead, GimmBO boosts visually similar alternatives to approach the target, but at $N{=}360$ still stalls short of a match.
This reflects the style-space observation of Sec.~\ref{sec:background}: visually similar adapters can produce similar images even when the recovered active set is wrong.

\subsubsection{Library scale}
\label{sec:eval-scale}

We next test how each method scales with library size using the nested libraries from Sec.~\ref{sec:eval-quant} (Fig.~\ref{fig:e2e}, right).
Using the nested libraries in Fig.~\ref{fig:e2e} (right), we increase $N$ from 40 to 1080 while keeping every target unchanged, so each larger library adds only distractor adapters, isolating sensitivity to library size from changes in the matching task.
At $N{=}40$, the scale GimmBO was designed for, HimmBO and GimmBO perform similarly, while Stylus+GimmBO coincides with GimmBO because retrieval returns the full library.
As $N$ grows, HimmBO maintains similar similarity, while Stylus+GimmBO degrades but remains the strongest alternative and GimmBO drops sharply.
Stylus stays nearly flat across library sizes, while random coordinate descent remains low.
Sparsity recovery explains these trends.
HimmBO recovers the most GT adapters at every size, although recovery declines for the two largest libraries as screening becomes harder.
Stylus+GimmBO again lies in between because retrieval removes many distractors, but any GT component it misses remains unavailable to GimmBO.
GimmBO's recovery falls to near zero as the library grows because its coefficient-space surrogate cannot transfer feedback between visually similar adapters, so additional adapters act mainly as distractors.
Despite HimmBO's lower recovery at larger $N$, its similarity drops much less because larger libraries also contain closer visual substitutes for missed GT components.
Refinement can exploit these substitutes, a property we develop further in Sec.~\ref{sec:eval-ext}.

\subsubsection{Ablations}
\label{sec:eval-ablation}

We ablate using our simulated matching task with $N=360$ for our screening and refinement stages respectively.

\paragraph{Screening.}
Following Sec.~\ref{sec:screening}, we compare UCB, ALM~\cite{mackay1992information,seo2000active}, and straddle as representative acquisition rules for BO, AL, and LSE, respectively (Fig.~\ref{fig:mechanism}).
At $N{=}360$, UCB improves fastest initially but plateaus as it concentrates on an early high-preference region.
ALM explores more broadly and improves more steadily.
Straddle spends its early rounds resolving uncertain level-set membership, then overtakes both before our twelve-round screening budget ends and achieves the highest recovery.
The dashed continuation beyond twelve rounds shows how these trends continue past the session budget.
At $N{=}40$, all three rules perform similarly because twelve rounds expose much of the library, but their differences grow with library size and straddle leads from $N{=}360$ onward.
This supports our use of LSE for screening: it spends several early rounds identifying promising regions, but becomes increasingly valuable once the library is too large to explore directly.

\begin{figure*}[tb]
  \centering
  \includegraphics[width=\linewidth]{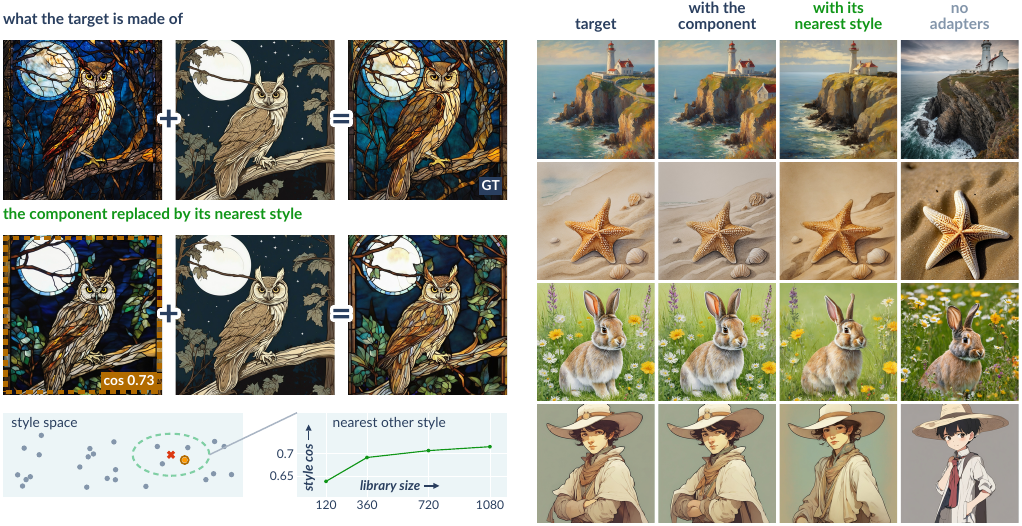}
  \caption{
  When a target component is missing from the library, HimmBO substitutes the nearest available style.
As the library grows, closer substitutes become available (left; style cos is the DreamSim cosine between the missing adapter and its nearest neighbor); for four additional targets, removing one GT component still lets HimmBO recover close approximations by substituting visually similar adapters (right).
}
  \label{fig:substitution}
\end{figure*}

\begin{figure*}[tb]
  \centering
  \includegraphics[width=\linewidth]{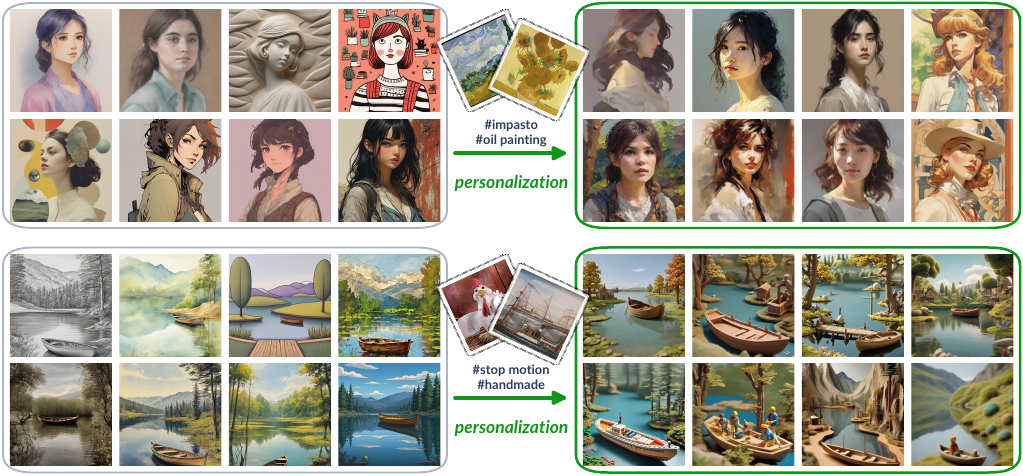}
  \caption{
  Reference images let HimmBO personalize screening before interaction begins.
They steer early screening from a broad coverage of the library (left) toward styles matching the user's preference, without restricting subsequent exploration (right); two examples are shown.
}
  \label{fig:personalize}
\end{figure*}

\begin{figure}[tb]
  \centering
  \includegraphics[width=\linewidth]{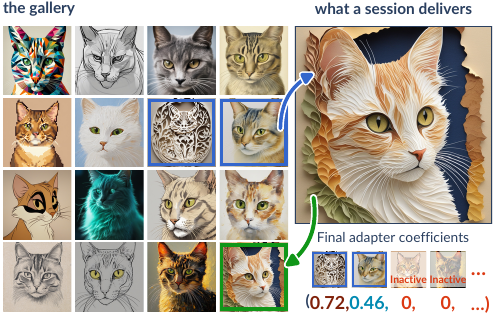}
  \caption{
  Each completed session can expand the screening pool.
Starting from single-adapter gallery samples (left), HimmBO delivers a preferred multi-adapter merge together with its coefficients and style feature (right), allowing it to be added directly without additional image generation.
}
  \label{fig:growth}
\end{figure}

\paragraph{Refinement.}
We ablate four refinement components while fixing the screening result so every variant starts from the same active set.
Removing the style kernel reduces similarity because feedback can no longer transfer between visually similar adapters.
Fixing the stick length to $B$, as in GimmBO, causes the largest loss by concentrating search near the cap rather than covering the broader range of total strengths seen in community merges.
Removing the diversity term has almost no effect on similarity, as intended, because it diversifies each batch rather than improving the optimum directly.
Without it, coefficient-space exploration can produce different merges with nearly identical styles, while the diversity term spreads them in style space (Fig.~\ref{fig:mechanism}, right).
Finally, discarding screening comparisons also reduces similarity, showing that refinement benefits from carrying earlier feedback into the reduced search space.

\subsection{Extensions}
\label{sec:eval-ext}

We explore three extensions of HimmBO.
Each builds on the default setup of Sec.~\ref{sec:eval-quant} and the style map's ability to encode arbitrary images, with all other session parameters unchanged.

\paragraph{Styles the library does not hold.}
The targets in Sec.~\ref{sec:eval-quant} are built from adapters in the library, but a user's desired style may have no exact match.
To test this, we remove one GT component from the library for each target and rerun the same session.
HimmBO still approximates the target by substituting a visually similar adapter for the missing component (Fig.~\ref{fig:substitution}).
How well this works depends on the substitute: when sufficiently similar, the final merge still reaches a rough match (DreamSim above $0.9$).
As the library grows, screening faces more distractors, but denser style coverage also makes closer substitutes more likely.

\paragraph{Personalization.}
A user may already have reference images that indicate their preferred style before interaction begins.
We use these images to initialize the preference model by averaging their representations into a single target for a simulated user and collecting three rounds of screening feedback.
The subsequent interaction is otherwise unchanged, except that the user completes only $T_1=9$ screening rounds instead of the default $12$.
The references therefore guide early discovery while leaving the full adapter library available for exploration (Fig.~\ref{fig:personalize}).
Screening then proceeds normally, without restricting the search to the neighborhood of the references.

\paragraph{A gallery that grows with use.}
We initialize the screening pool with single-adapter samples because they reliably produce valid, high-quality images.
Generic space-filling sampling over multi-adapter coefficients can instead produce many poor-quality combinations, as observed by previous work~\cite{liu2026gimmbo}.
Refinement provides a better source of multi-adapter samples: every delivered merge has already produced a valid image and been preferred by a user.
Because its coefficients and style feature are also known, we can add it directly to the screening pool without training a new adapter or generating another image (Fig.~\ref{fig:growth}).
In a preliminary experiment, adding such merges increases the mean final similarity from $0.923$ to $0.944$ (App.~\ref{sec:supp-growth}).
Over time, these user-vetted multi-adapter samples could improve future screening with little additional acquisition overhead.

%% file: sections/05_conclusion.tex
\section{Limitations and Future Work}
\label{sec:conclusion}

Our evaluation uses linear merging only (Eq.~\eqref{eq:merge}).
Screening itself is independent of the merging rule because its gallery contains only single-adapter images, so alternative rules~\cite{matena2021fisher,yadav2023ties,yu2024dare} would affect refinement alone and are straightforward to explore in future work.

Our evaluation relies on simulated preference feedback.
GimmBO's user study found DreamSim to be a good proxy for human judgments in the same matching task~\cite{liu2026gimmbo}, supporting its use for controlled comparisons here.
We additionally conducted a small preliminary test with real users while measuring session cost (App.~\ref{sec:supp-cost}); across these multi-round sessions, HimmBO found better merges than GimmBO while completing the interaction faster.
Our simulated user, however, provides noise-free and internally consistent feedback, whereas real users may be inconsistent or change their preferences during a session.
A larger user study is needed to measure how often this occurs and whether it causes screening to select the wrong adapters or refinement to become less stable.
These effects could motivate preference models that are more robust to noisy or drifting feedback.

Our adaptation uses only part of the standard LSE formulation.
Standard LSE assumes scalar Gaussian observations and intersects confidence intervals across rounds to bound a fixed latent function on a common scale~\cite{gotovos2013active}.
Our utilities are instead inferred from comparisons and the posterior is refit each round, so confidence intervals from different rounds are not directly comparable.
We therefore use straddle and the per-round level-set partition, but omit interval intersection and related components.
Extending these ideas to preferential feedback, including more recent level-set methods, is an interesting direction for future work.

The style map suppresses content shared across gallery images but does not remove it completely, so a reference image with different content may retrieve the wrong style neighbors.
A representation that separates style and content more explicitly could make personalization more reliable across subjects.
The transition from screening to refinement is also one-way, so an adapter omitted from the active set cannot be recovered later.
Revisiting the full library as refinement progresses could correct such misses and make the two stages interact more tightly.
Finally, a gallery that retains delivered merges (Sec.~\ref{sec:eval-ext}) grows indefinitely.
Developing policies for which samples to retain or retire could preserve the benefits of gallery growth while keeping acquisition efficient.

Midjourney's Style Creator\footnote{\url{https://docs.midjourney.com/hc/en-us/articles/41308374558221-Style-Creator}} offers related capabilities, but represents the resulting style as a position in its proprietary style space.
Our approach instead searches an extensible library of adapters, returning the adapters and coefficients that compose the result while allowing the library itself to grow over time.

Adapters also inherit and can amplify the harms of generative image models.
A large part of what Civitai hosts exists to sexualize women, and a session can start from any photograph, so the same search machinery can be applied to a real person who never consented to it.
Searching a whole library can also make it easier to find and combine adapters that imitate a living artist, which artists are actively trying to prevent~\cite{shan2023glaze}.
We plan to release our library as a manifest and curation pipeline rather than redistribute adapter weights.
This does not resolve these broader problems, but safer curation and access mechanisms for large model collections are important directions for future work.

%% file: sections/06_supplemental.tex
\clearpage
\setcounter{figure}{0}\renewcommand{\thefigure}{S\arabic{figure}}
\setcounter{table}{0}\renewcommand{\thetable}{S\arabic{table}}
\setcounter{equation}{0}\renewcommand{\theequation}{S\arabic{equation}}

\section*{Supplemental Material}

\appendix
\setcounter{secnumdepth}{2}

\section{Implementation details}
\label{sec:supp-impl}

\subsection{Surrogate hyperparameters and priors}
\label{sec:supp-hyper}

The kernel in Eq.~\eqref{eq:kernel} has four hyperparameters: the channel variances $\sigma_s^{2},\sigma_I^{2}$ and length scales $\ell_s,\ell_I$, with one length scale shared across each channel.
We re-fit these hyperparameters each round by maximum a posteriori estimation using all comparisons collected so far.
We fix the probit noise to one because the likelihood depends only on the ratio between utility differences and noise, and this scale is already controlled by the channel variances.
Each hyperparameter has a log-normal prior centered at a value calibrated offline on simulated sessions separate from the evaluation, with standard deviation $0.5$ in log space; nearby values perform similarly.
This prior is important because, when comparisons are largely consistent, the likelihood continues to increase as the latent utilities are scaled up and therefore has no finite maximum.
Without the prior, the hyperparameters drift to extreme values and search performance degrades, while fixing them at their calibrated values also performs worse.
The prior center for $\sigma_s^{2}$ is larger during screening than refinement (Table~\ref{tab:supp-constants}), likely because screening searches the full library rather than the $m$ active adapters and therefore benefits from stronger style guidance.

Unlike GimmBO, we do not use a sparsity-inducing prior such as SAAS~\cite{eriksson2021saas}, which assigns a separate length scale to each input dimension and shrinks most dimensions.
Our style representation is already compact after PCA, and we instead use a single shared length scale per channel rather than selecting individual feature dimensions.
This also reduces each hyperparameter re-fit from about ten seconds to under one second.

In refinement, we use a smaller exploration weight than GimmBO ($\lambda=1$ versus $9$) and add the diversity term with weight $\gamma$.
Because screening has already explored the full library, refinement searches only merges of the $m$ active adapters and requires less exploration.
Empirically, the smaller exploration weight also improves performance.
All constants are listed in Table~\ref{tab:supp-constants}.

\begin{table}[t]
  \centering
  \small
  \caption{HimmBO's hyperparameters. Kernel hyperparameters are re-fit every round (Sec.~\ref{sec:supp-hyper}), so the table reports their prior centers.}
  \label{tab:supp-constants}
  \begin{tabular}{@{}lp{1.72in}r@{}}
    \toprule
    Symbol & Meaning & Value \\
    \midrule
    $q$ & images per batch & 8 \\
    $T_1$ & screening rounds & 12 \\
    $T_2$ & refinement rounds & 8 \\
    $m$ & size of the active set & 8 \\
    $B$ & total-strength cap, Eq.~\eqref{eq:domain} & $2$ \\
    $\tau$ & per-adapter floor, Eq.~\eqref{eq:domain} & $0.1$ \\
    $\theta$ & quantile defining the level $\hat h$, Eq.~\eqref{eq:straddle} & $0.5$ \\
    $\rho_\ell$ & $\ell_s$ as a multiple of the median pairwise style distance & $0.5$ \\
    $\sigma_s^{2}$ & style-channel variance, Eq.~\eqref{eq:kernel}; screening / refinement & $16$ / $4$ \\
    $\sigma_I^{2}$ & coefficient-channel variance, Eq.~\eqref{eq:kernel} & $0.5$ \\
    $\ell_I$ & coefficient-channel length scale, Eq.~\eqref{eq:kernel} & $0.5$ \\
    $\lambda$ & UCB exploration weight & $1$ \\
    $\gamma$ & diversity weight, Eq.~\eqref{eq:acq} & $0.02$ \\
    $t_0$ & SDEdit strength & $0.8$ \\
    $\ldiv$ & length scale of the diversity kernel & $0.5$ \\
    $\varepsilon$ & jitter in the log-determinant & $10^{-3}$ \\
    \bottomrule
  \end{tabular}
\end{table}

\subsection{Offline precomputation}
\label{sec:supp-offline}

\paragraph{The gallery.}
\label{sec:supp-gallery}
Each adapter is generated alone at unit strength ($\alpha_i=1$) on four fixed prompts, covering a person, an animal, a still life, and a landscape, so that each style is observed across different subject types.
Because all adapters use the same prompts, differences across their galleries primarily reflect the adapters' effects.
For our experiments, we use single adapters rather than merges because platforms already publish and preview adapters this way.
Sec.~\ref{sec:eval-ext} explores extending the gallery with user-vetted multi-adapter merges collected during refinement.

The style map below maps each gallery image to a style coordinate.
For a given session prompt, adapters are compared using the coordinates of their corresponding gallery images.
We denote the coordinate of adapter $i$ by $\z_i$.
The gallery stores the screening images and, for each prompt, an $N\times d$ coordinate matrix used by the posterior.
The images dominate storage, while all session-time computation uses only these matrices.

Building the gallery requires four inferences per adapter, paid once per library: about 4 GPU-hours at $N{=}1080$, shared by all sessions and users.
Adding a new adapter requires only generating its gallery images and mapping them through the existing style map.
The style kernel matrix then gains one row and column, and the median used to set $\ell_s$ is recomputed in seconds.

\begin{table}[t]
  \centering
  \small
  \caption{Final similarity under different PCA variance thresholds, in the main-experiment setting ($N{=}360$).}
  \label{tab:supp-pca}
  \begin{tabular}{@{}lccccc@{}}
    \toprule
    Explained variance & 0.5 & 0.6 & 0.7 & \textbf{0.8} & 0.9 \\
    \midrule
    $d$ & 20 & 32 & 53 & \textbf{90} & 159 \\
    Similarity & 0.893 & 0.910 & 0.901 & \textbf{0.925} & 0.917 \\
    \bottomrule
  \end{tabular}
\end{table}

\paragraph{The style map.}
\label{sec:supp-stylemap}

Because all adapters share the same four prompts, images generated from a given prompt share much of their content.
The style map $\Phi$ centers the features within each prompt, normalizes them, and reduces their dimension with PCA,
\begin{equation}
  \z_i^{p}=\Phi\big(I_i^{p}\big)=\PCA\Big(\nrm\big(\varphi(I_i^{p})-\tfrac{1}{N}\textstyle\sum_j\varphi(I_j^{p})\big)\Big),
  \label{eq:stylemap}
\end{equation}
where $\varphi$ is a perceptual image encoder, DreamSim by default (other encoders are compared in App.~\ref{sec:supp-encoder}).
For each test case, we use the gallery prompt whose subject type best matches the test prompt (e.g., the animal prompt for an animal target) and use its style coordinates throughout the session.

Centering subtracts the mean feature for each prompt, which primarily captures shared subject and composition, emphasizing the adapters' contributions.
This step is essential: without it, nearest-neighbor cosine similarities are all close to one, and the distance-based components of our method (the style kernel, exclusion radius, and peak readout) can no longer distinguish adapters.
Normalization projects each feature onto the unit sphere so that distances reflect style rather than feature magnitude.
PCA then retains enough components to explain $80\%$ of the library variance, giving $d{=}90$ at $N{=}360$.
Retaining more dimensions provides no gain, while retaining fewer reduces similarity (Table~\ref{tab:supp-pca}).

We center the prior for the style length scale relative to the library,
$\ell_s=\rho_\ell\med_{i\neq j}\norm{\z_i-\z_j}$ (Table~\ref{tab:supp-constants}),
so it does not require retuning as the library changes.

\subsection{Screening}
\label{sec:supp-lse}

\paragraph{Adapting LSE.}
Following LSE~\cite{gotovos2013active}, we partition adapters into three sets each round using their posterior intervals.
Let $Q_t(i)=[\mu_{t,i}-\beta\sigma_{t,i},\,\mu_{t,i}+\beta\sigma_{t,i}]$ denote the interval for adapter $i$ at round $t$, with $\beta=1.96$ as in Eq.~\eqref{eq:straddle}.
Adapters whose intervals lie entirely above or below the level $\hat h_t$ form
\begin{equation}
  H_t=\{i:\mu_{t,i}-\beta\sigma_{t,i}\ge\hat h_t\},\qquad
  L_t=\{i:\mu_{t,i}+\beta\sigma_{t,i}\le\hat h_t\},
  \label{eq:hlu}
\end{equation}
and the remaining adapters, whose intervals contain $\hat h_t$, form the uncertain set $U_t$.
The straddle score in Eq.~\eqref{eq:straddle} implements this partition directly: $a_t(i)>0$ exactly when $\beta\sigma_{t,i}>\lvert\mu_{t,i}-\hat h_t\rvert$, so we select each batch only from $U_t$.
Early in a session, unexplored adapters have posterior means near the level and large uncertainty, placing them in $U_t$ and causing them to be queried early.
Because the three sets are rebuilt from the current posterior each round, an adapter can leave $H_t$ or $L_t$ as later comparisons shift its posterior mean, allowing earlier classifications to be revised as more feedback arrives.

We omit the remaining components of the standard LSE formulation~\cite{gotovos2013active}, whose assumptions do not carry over to preferential feedback.
Standard LSE observes noisy scalar values of a fixed function $f$ under a Gaussian likelihood with fixed kernel hyperparameters, yielding an exact posterior whose intervals across rounds bound the same $f$ on a common scale.
Our posterior is instead inferred from comparisons through a Laplace approximation, with hyperparameters re-fit every round, so its utility scale can change across rounds.
Standard LSE intersects intervals across rounds and schedules $\beta$ so that they jointly cover $f$.
Because our intervals are not comparable across rounds, we keep $\beta$ fixed, do not intersect intervals, and allow classified adapters to be reconsidered.
Its batch variant also spreads queries by updating the variance as if pending observations were already available, which relies on Gaussian observations.
We instead separate each batch using the style radius of Sec.~\ref{sec:screening}.
Finally, standard LSE stops once the classification is $\varepsilon$-accurate, whereas we stop after a fixed budget of $T_1$ rounds to bound user interaction time.

\begin{figure}[t]
  \centering
  \includegraphics[width=\linewidth]{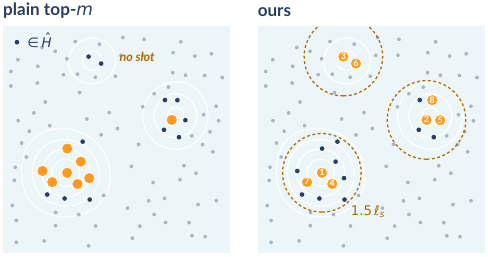}
  \caption{Plain top-$m$ spends most of the $m{=}8$ slots on the highest peak and leaves one peak without a slot (left), while the peak readout shares the slots across peaks (right).
Contours show the posterior mean, and numbers give the order in which the slots are filled.}
  \label{fig:supp-readout}
\end{figure}

\paragraph{Determining the active set.}
\label{sec:supp-readout}
After $T_1$ rounds, a few hundred comparisons over $N$ adapters leave substantial posterior uncertainty for many candidates.
Because $H_{T_1}$ includes only adapters whose entire posterior interval lies above the level, it contains only those already classified with high confidence.
We therefore read out the active set from the estimated level set $\hat H$ of Sec.~\ref{sec:screening}, which also includes adapters whose posterior mean is above the level but whose membership remains uncertain.
Because preference often has several peaks in style space (Sec.~\ref{sec:background}), we allocate the $m$ slots across peaks so that the active set represents multiple promising regions rather than containing $m$ lookalikes from the highest one (Fig.~\ref{fig:supp-readout}).

We identify peaks greedily by visiting the adapters in $\hat H$ in decreasing order of posterior mean (Fig.~\ref{fig:supp-readout}).
The first adapter initializes a peak, and each later adapter initializes a new peak if it lies more than $1.5\,\ell_s$ from all existing peak founders.
Otherwise, it joins an existing peak whose founder lies within $\ell_s$.
We then fill the active set by taking the best unselected member from each peak in turn, following the order in which the peaks were created, until all $m$ slots are filled.
If the peaks run out of members, we fill the remaining slots from $\hat H$ in decreasing order of posterior mean, although $\hat H$ typically contains far more than $m$ adapters.
With only one peak, this procedure reduces to plain top-$m$ selection.

The peak readout raises the F1 of the active set by about $0.04$ over plain top-$m$.
The separation radius of $1.5\,\ell_s$ gives the highest F1, smaller radii stay within $0.01$ of it, and larger radii merge distinct peaks and fall toward plain top-$m$ (Table~\ref{tab:supp-sep}).

\begin{table}[t]
  \centering
  \small
  \caption{F1 of the active set under different separation radii (10 seeds).}
  \label{tab:supp-sep}
  \begin{tabular}{@{}lccccccc@{}}
    \toprule
    Radius$/\ell_s$ & 0.75 & 1.0 & 1.25 & \textbf{1.5} & 1.75 & 2.0 & 2.5 \\
    \midrule
    F1 & 0.369 & 0.370 & 0.370 & \textbf{0.374} & 0.372 & 0.358 & 0.329 \\
    \bottomrule
  \end{tabular}
\end{table}

\subsection{Refinement}
\label{sec:supp-acqopt}

\paragraph{Predicting the style of a merge.}
Because acquisition evaluates a proposed merge before its image is generated, its style feature is not yet available.
We therefore predict it from the gallery as the coefficient-weighted sum of the active adapters' style features,
\begin{equation}
  \zhat(\balpha)=\sum_{i\in\Aset}\alpha_i\,\z_i .
  \label{eq:zhat}
\end{equation}
This prediction is used only for acquisition: once a merge is generated, it enters the posterior with its measured $\z$, so prediction errors affect which merges are proposed but not what the model learns.
Because $\zhat$ is linear in $\balpha$, Eq.~\eqref{eq:acq} remains differentiable end to end.
As in GimmBO, we optimize it with multi-start L-BFGS initialized from stick-breaking samples of the domain.

\begin{figure}[t]
  \centering
  \includegraphics[width=\linewidth]{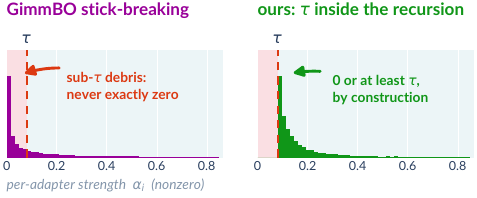}
  \caption{Nonzero per-adapter strengths: thresholding GimmBO's map afterwards leaves sub-$\tau$ residuals (left), while enforcing $\tau$ inside the recursion does not (right).}
  \label{fig:stickbreak-tau}
\end{figure}

\paragraph{Sampling the domain.}
GimmBO maps the unit cube onto the capped simplex using stick-breaking.
Starting from $R_0=B$, each coordinate takes a fraction of the remaining stick, $\alpha_i=\min(x_iR_{i-1},1)$ and $R_i=R_{i-1}-\alpha_i$, with a random coordinate order to avoid favoring any axis.
Every sample respects the cap, and most coordinates remain small, encouraging sparse merges.
However, the total strength is strongly biased toward the cap: for uniform inputs, $\E[\sum_i\alpha_i]=B(1-2^{-m})$, which is within $1\%$ of $B$ at $m=8$, whereas real community merges span a much broader range of lower totals (inset in Sec.~\ref{sec:refine}).

We therefore replace the fixed $R_0=B$ with a variable stick length $S\le B$.
When drawing samples for acquisition optimization, we fit the distribution of $S$ to the total strengths of over a thousand popular community merges, so the samples used to seed the L-BFGS restarts have realistic totals.
During optimization, $S$ becomes the $(m{+}1)$-th decision variable, allowing the search to optimize total strength directly rather than remain near the cap.
The empirical distribution therefore affects only the restart samples, not the feasible domain, so totals rarely used by the community remain reachable during optimization.

Thresholding the stick-breaking output against $\tau$ after the fact either violates the cap or leaves sub-$\tau$ residuals, coefficients too small to matter but still nonzero (Fig.~\ref{fig:stickbreak-tau}, left).
We instead enforce $\tau$ directly within the recursion.
With $R_0=S$ and $\bar u_i=\min(1,R_{i-1})$,
\begin{equation}
  \alpha_i=
  \begin{cases}
    0 & \text{if } \bar u_i<\tau,\\
    \tau+x_i\,(\bar u_i-\tau) & \text{otherwise},
  \end{cases}
  \qquad
  R_i=R_{i-1}-\alpha_i,
  \label{eq:tauin}
\end{equation}
so each break is either zero or at least $\tau$, and every sample lies exactly in $\dom$ (Fig.~\ref{fig:stickbreak-tau}, right).

\subsection{Image generation}
\label{sec:supp-imagegen}
All images are generated with the SDXL 1.0 base model~\cite{podell2024sdxl} using the diffusers pipeline in half precision at a resolution of $1024\times1280$.
Following GimmBO~\cite{liu2026gimmbo}, we use the DPM++ 2M sampler with the Karras noise schedule, $30$ denoising steps, and a classifier-free guidance scale of $7$.
During refinement, the $m$ selected adapters are loaded once, and each merge fuses their weight updates into the base model using the coefficients $\balpha$ from Eq.~\eqref{eq:merge} before image generation.
Gallery images use seed $42$ by default and are manually checked for quality, while all other images use the same five fixed seeds across methods, one per run.

Gallery images are generated from text alone, so that each adapter's style is preserved without content constraints.
During a session, we control content with SDEdit~\cite{meng2022sdedit}.
Each session first generates an anchor image from its prompt with the base model alone, and every merge is then generated by noising this anchor to $t_0=0.8$ and denoising from there, which runs $24$ of the $30$ steps.

\subsection{Cost and engineering}
\label{sec:supp-cost}

Search computation remains small at every library size.
Screening requires no GPU: each round consists of one posterior fit and one scan of the library, taking about $0.3$\,s.
Gallery images that have not appeared in any comparison contribute nothing to the likelihood and can be integrated out exactly under the Gaussian process prior, so we fit the posterior only on compared images.
During refinement, when only a few new images are added each round, this is about ten times cheaper than fitting on the entire gallery.
The remaining per-round cost, scoring all adapters, scales linearly with $N$.
The two quadratic quantities, the style kernel matrix and the median distance used to set $\ell_s$, are computed once when the library is built and updated only when the library changes (Sec.~\ref{sec:supp-gallery}).
Refinement always receives $m=8$ adapters, which are loaded once at its start, so its session start-up cost is independent of $N$.

\begin{figure}[t]
  \centering
  \includegraphics[width=\linewidth]{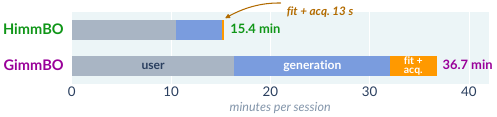}
  \caption{Wall-clock time of one session, from the per-round times of Table~\ref{tab:supp-timing}.
  HimmBO requires less user interaction time because its screening rounds are faster (Table~\ref{tab:utime}).}
  \label{fig:supp-sessiontime}
\end{figure}

We report wall-clock time per round (Table~\ref{tab:supp-timing}) and per session (Fig.~\ref{fig:supp-sessiontime}) on a single H100 GPU with the $N{=}360$ library, averaged over ten sessions done by two novice users.
During refinement, each round generates $q$ merges, while posterior updating and acquisition optimization take about $1$\,s compared with about $35$\,s for image generation.
Thus, in both stages, user-facing latency is dominated by image generation or the user's own judgment rather than search computation.
On average, generation takes $4.4$\,s per image in HimmBO and $5.7$\,s in GimmBO because HimmBO refines only over the $m{=}8$ selected adapters, which remain in memory, whereas GimmBO searches the full library and often needs to load adapters from storage.
A HimmBO session takes about $15$ minutes, compared with $37$ minutes for GimmBO.
Most of this difference comes from screening, whose $12$ rounds require no image generation; the remainder comes from shorter user judgment per round and faster model fitting, which takes about ten seconds per round for GimmBO under its SAAS prior (Sec.~\ref{sec:supp-hyper}).
Participants who used both methods reported that HimmBO performs better than GimmBO.

\begin{table}[t]
  \centering
  \small
  \caption{Wall-clock seconds per round on a single H100 GPU with the $N{=}360$ library.}
  \label{tab:supp-timing}
  \setlength{\tabcolsep}{4pt}
  \begin{tabular}{@{}lrrrrr@{}}
    \toprule
    & user & gen. & fit & acq. & total \\
    \midrule
    HimmBO screening & 24.4 & 0 & 0.28 & 0.02 & 24.7 \\
    HimmBO refinement & 41.9 & 35.2 & 0.21 & 0.97 & 78.3 \\
    GimmBO & 46.8 & 46.3 & 10.8 & 2.6 & 106.5 \\
    \bottomrule
  \end{tabular}
  \label{tab:utime}
\end{table}

\section{Experimental setup}
\label{sec:supp-setup}

\subsection{The library}
\label{sec:supp-library}

\begin{table}[t]
  \centering
  \small
  \caption{Adapters left after each curation step.}
  \label{tab:supp-funnel}
  \begin{tabular}{@{}lr@{}}
    \toprule
    Step & Adapters \\
    \midrule
    Crawl: SDXL~1.0 adapters tagged \emph{style} & 9{,}909 \\
    Filter: downloads and tags & 1{,}832 \\
    Language-model check & 1{,}359 \\
    Style de-duplication & 1{,}200 \\
    Generated successfully & 1{,}172 \\
    Manual pass & 1{,}080  \\
    \bottomrule
  \end{tabular}
\end{table}

\paragraph{Curation.}
On August 18, 2026, we queried the public Civitai API for LoRA adapters tagged \emph{style} with SDXL~1.0 as the base model, obtaining 9{,}909 adapters.
Table~\ref{tab:supp-funnel} lists how many adapters remain after each curation step of Sec.~\ref{sec:eval-quant}.
Adapters for derived base models such as Pony and Illustrious rarely work on SDXL and are excluded.
We keep adapters with at least 1{,}000 downloads and drop those whose tags or names indicate content other than style, such as characters, clothing, poses, detail enhancers, and sliders, as well as those lacking a downloadable SDXL version or a preview image.
We also retain all adapters from GimmBO's library that are still online.
Because tags cannot distinguish a style from an adapter that adds content, a language model (Claude Haiku) reads each remaining adapter's name, tags, trigger words, and description, keeping only those that change how an image looks rather than what it shows.
Since independently trained adapters can produce nearly identical looks (Fig.~\ref{fig:stylecoef}), we repeatedly remove the less downloaded adapter of the most similar pair, measured by CSD features~\cite{somepalli2024csd} of their preview images, until 1{,}200 adapters remain.
We then generate each adapter alone at unit strength from a fixed seed and a short prompt beginning with its trigger word.
Of these, 1{,}172 produce an image, while the rest are no longer available or use unsupported formats.
Finally, we inspect all generated images and remove adapters that do not act as styles, mostly those that replace the subject, such as building blocks or game icons, or add a generic look, such as \emph{cinematic}.

\paragraph{Nested libraries.}
The full library, \emph{SDXL-Style-1K}, contains 1{,}080 adapters with a median of 2{,}100 downloads, including GimmBO's 40 adapters.
We select the $N{=}360$ library to cover the style space broadly, then remove or add adapters to obtain the $N{=}40$, 120, and 720 libraries, so that all five are nested, $40\subset120\subset360\subset720\subset1080$.
We will release the manifest of all five libraries, including each adapter's Civitai identifier and version.

\subsection{Targets and simulated user}
\label{sec:supp-targets}
\label{sec:supp-simuser}

The $24$ ground-truth (GT) targets cover the four subject types used by the gallery prompts (a person, an animal, a still life, and a landscape) with six targets each.
Their prompts will be released with the code.
Half of the GT targets are generated from hidden merges of two library adapters and half from three, the most common merge sizes in community use and small enough for each component's style to remain distinguishable.
Their total strengths are drawn from the empirical distribution of community merges, using the same usage statistics as in Sec.~\ref{sec:supp-acqopt}.
To avoid low-quality targets, we manually inspect the sampled merges and exclude cases where a component's style is barely visible or the generated image is low quality.

The simulated user follows GimmBO's protocol (Sec.~\ref{sec:eval-quant}), with its choices converted into pairwise comparisons as follows.
During screening, each of the $c$ marked images is preferred over each of the $q-c$ unmarked images, yielding $16$ comparisons per round.
During refinement, each of the top five images is preferred over every image ranked below it or left unranked.

\subsection{Baselines}
\label{sec:supp-baselines}
GimmBO runs with its published implementation and configuration~\cite{liu2026gimmbo}.

For Stylus~\cite{luo2024stylus}, we rebuild the retrieval index on our library rather than using Stylus's own adapter index.
A vision-language model (Gemini 3.7 Flash) reads each adapter's model card and three of its images, one from our gallery, and writes a description of the adapter.
We embed these descriptions with bge-large-en-v1.5~\cite{xiao2024cpack} and retrieve the $150$ adapters closest to the query by cosine similarity, where the query is a description of the target's components written by the same model.
Stylus's composer and masking steps are kept as published, with the same model as the composer.

In our simulated-user experiments, across rounds, we rerun Stylus with a new seed, each round taking the top $8$ adapters not taken in earlier rounds, then keep the running best solution.
Stylus+GimmBO instead runs GimmBO on the top $40$ of the $150$ retrieved adapters after running one Stylus retrieval at the beginning.

Random coordinate descent follows GimmBO's slider baseline.
It visits the library's adapters in random order and sweeps each one's coefficient over eight evenly spaced values in $[0,1]$ while keeping the others fixed, and the simulated user keeps the most similar image as the new incumbent.
The range includes $0$ so that the baseline can switch adapters off, and we do not impose the cap $B$.
Images already generated are reused rather than regenerated.

\section{Choice of perceptual encoder}
\label{sec:supp-encoder}

The style map can use any perceptual encoder $\varphi$ (Sec.~\ref{sec:style}).
We compare DreamSim against three alternatives: the recent text-prompted perceptual metric TPIPS~\cite{wang2026tpips}, prompted with \emph{style}; CLIP ViT-L/14~\cite{radford2021clip}; and the style descriptor CSD~\cite{somepalli2024csd}.
For each run, we replace the encoder in both the style map and the simulated user while keeping all other settings of the main $N{=}360$ experiments unchanged.
Each row of Table~\ref{tab:supp-encoder} therefore evaluates a complete pipeline built on one encoder, rather than isolating the style kernel alone.
We then score the final merge from each run against the GT using all four encoders.

The DreamSim pipeline achieves the highest score under three of the four evaluation encoders, while the TPIPS pipeline is slightly higher under TPIPS ($0.923$ versus $0.921$).
Notably, when scored with CLIP or CSD, the DreamSim pipeline also outperforms the pipelines that use CLIP or CSD during search.
Thus, its advantage is not confined to evaluation with the same encoder used during search.
In the DreamSim column, all three alternative pipelines also outperform GimmBO ($0.808$, Sec.~\ref{sec:eval-quant}), even though none uses DreamSim during search, whereas GimmBO's simulated user does.

\begin{table}[t]
  \centering
  \small
  \caption{Similarity of the final merge to the GT on the $N{=}360$ library.
Rows give the encoder used throughout the search, including the simulated user, and columns the encoder used for scoring; the best entry of each column is in bold.}
  \label{tab:supp-encoder}
  \begin{tabular}{@{}lcccc@{}}
    \toprule
    & \multicolumn{4}{c}{Scored with} \\
    \cmidrule(l){2-5}
    Searched with & DreamSim & TPIPS & CLIP & CSD \\
    \midrule
    DreamSim & \textbf{0.925} & 0.921 & \textbf{0.919} & \textbf{0.888} \\
    TPIPS & 0.892 & \textbf{0.923} & 0.904 & 0.860 \\
    CLIP & 0.875 & 0.901 & 0.911 & 0.853 \\
    CSD & 0.873 & 0.898 & 0.908 & 0.867 \\
    \bottomrule
  \end{tabular}
\end{table}

\section{How much feedback each kernel needs}
\label{sec:supp-quant}
\label{sec:supp-reach}

The inset of Sec.~\ref{sec:surrogate} measures the number of feedback rounds required before the active set recovers at least half of the target's components, using the nested libraries of Sec.~\ref{sec:eval-quant}.
The two curves follow the same protocol and differ only in the kernel: the full kernel of Eq.~\eqref{eq:kernel} or its coefficient channel alone.
We use pure posterior variance (ALM~\cite{mackay1992information}) for acquisition so that neither kernel is paired with an acquisition rule tailored to it.
Each round shows $q{=}8$ gallery images, of which the simulated user marks the $c{=}4$ closest to the target (Sec.~\ref{sec:supp-simuser}), yielding $16$ comparisons.
Thus, the ratio between the curves is identical whether feedback cost is measured in rounds or comparisons.
Adapters may be queried again in later rounds, and we impose no round limit because the coefficient-only kernel requires far more feedback than a normal session allows.
We read out the active set as in Sec.~\ref{sec:supp-readout} with $m{=}8$ and evaluate four targets over five seeds.

\section{A gallery that grows with use}
\label{sec:supp-growth}

We test whether adding delivered merges to the gallery (Sec.~\ref{sec:eval-ext}, Fig.~\ref{fig:growth}) improves later sessions.
To mimic popular adapters that many users combine, we take $12$ widely downloaded adapters with distinct styles from the $N{=}360$ library and sample $60$ merges of two or three of them, following the target protocol of Sec.~\ref{sec:supp-targets}.
We run the default session (Sec.~\ref{sec:eval-quant}) with each merge as the target.

We then select six GT targets, three with two components and three with three, whose original sessions achieve a rough match but not a near-exact one.
We add the remaining $54$ merges to the gallery, together with their coefficients, images, and style features.
Each added merge shares at least one component with some target, but none uses exactly the same component set as any target, so the gallery contains related combinations rather than the answers themselves.
An added merge enters the posterior like any other gallery sample: if selected by the procedure of Sec.~\ref{sec:supp-readout}, it occupies one slot while introducing all of its components into the active set.
We rerun the six targets with this expanded gallery and compare against their sessions on the original gallery averaged over the same five seeds.

The added merges improve active-set recovery for half of the targets and increase the mean final similarity from $0.923$ to $0.944$.
In practice, popular community merges and merges delivered in earlier sessions could be added in the same way without additional image generation, allowing the gallery to become denser in regions that users frequently prefer.
These results are preliminary, and confirming the effect at scale will require a larger study using real usage logs.

\section{Additional qualitative results}
\label{sec:supp-qual}

\begin{figure*}[b]
  \centering
  \includegraphics[width=\linewidth]{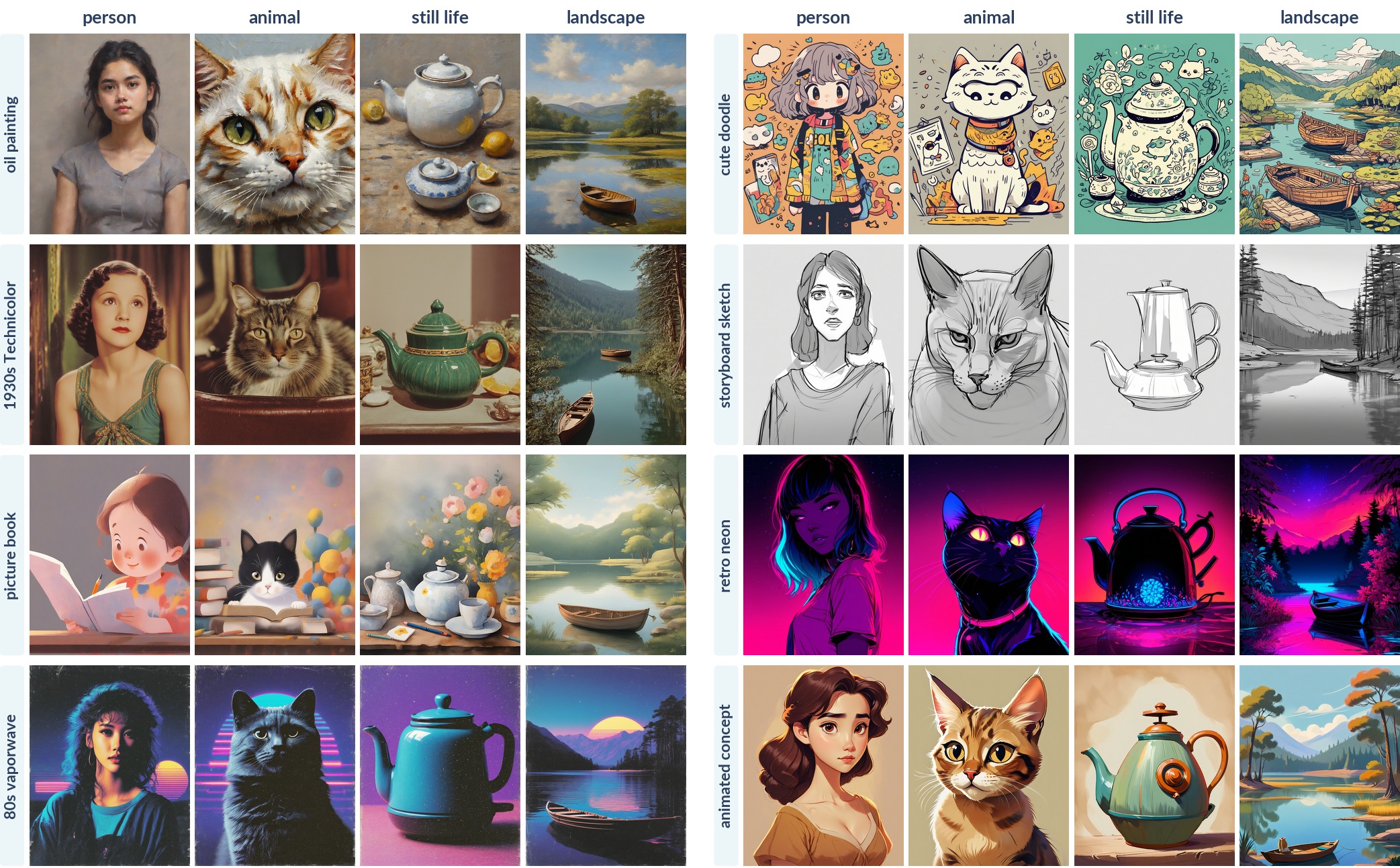}
  \caption{Eight adapters of the library, each generated alone at unit strength on the four gallery prompts.}
  \label{fig:supp-gallery}
\end{figure*}

Fig.~\ref{fig:supp-gallery} shows eight adapters of our library through their gallery images.

\begin{figure*}[t]
  \centering
  \includegraphics[width=\linewidth]{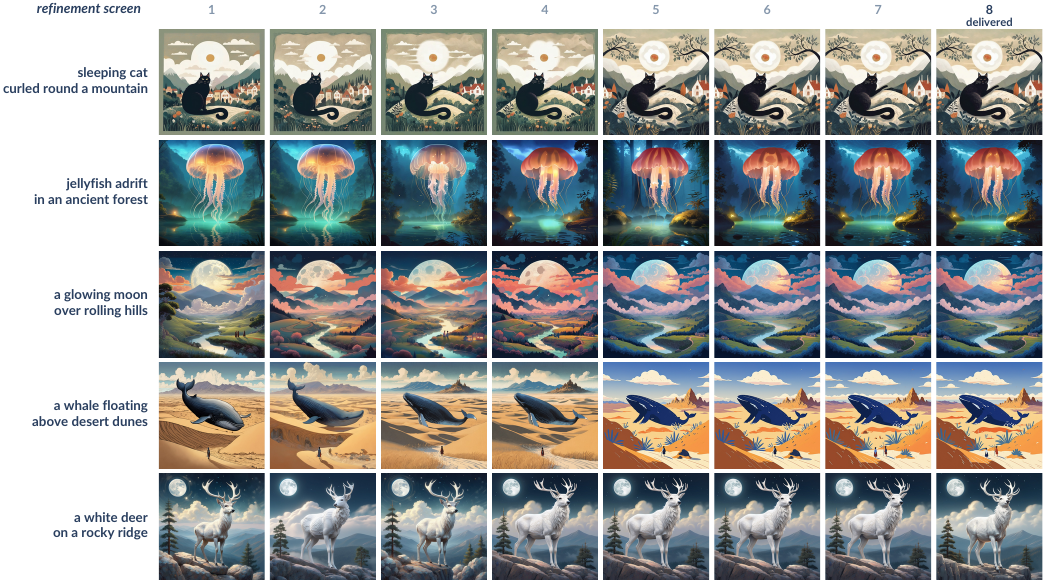}
  \caption{The refinement stage of five open-ended sessions, one per row.}
  \label{fig:supp-sessions}
\end{figure*}

Fig.~\ref{fig:supp-sessions} shows five \emph{open-ended} sessions with no target to match.
After screening the library as usual, participants write their own prompts and rank merges by preference alone.

